%% file: main.tex
\documentclass[10pt,letterpaper]{article}
\usepackage[top=0.85in,left=2.75in,footskip=0.75in]{geometry}

\usepackage{amsmath,amssymb}

\usepackage{changepage}

\usepackage[utf8x]{inputenc}

\usepackage{textcomp,marvosym}

\usepackage{cite}

\usepackage{booktabs}

\usepackage{nameref,hyperref}

\usepackage[right]{lineno}

\usepackage{microtype}
\DisableLigatures[f]{encoding = *, family = * }

\usepackage[table]{xcolor}

\usepackage{array}
\usepackage{enumitem}
\usepackage{graphicx}
\usepackage{subcaption}

\newcolumntype{+}{!{\vrule width 2pt}}

\newlength\savedwidth

\newcommand\thickhline{\noalign{\global\savedwidth\arrayrulewidth\global\arrayrulewidth 2pt}%
\hline
\noalign{\global\arrayrulewidth\savedwidth}}

\raggedright
\usepackage[aboveskip=1pt,labelfont=bf,labelsep=period,justification=raggedright,singlelinecheck=off]{caption}

\makeatletter
\renewcommand{\@biblabel}[1]{\quad#1.}
\makeatother

\usepackage{lastpage,fancyhdr,graphicx}
\usepackage{epstopdf}
\fancyheadoffset[L]{2.25in}
\fancyfootoffset[L]{2.25in}
\begin{document}
\vspace*{0.2in}

\begin{flushleft}
{\Large
\textbf\newline{Aggregating Soft Labels from Crowd Annotations Improves Uncertainty Estimation Under Distribution Shift} %
}
\newline
\\
Dustin Wright*,
Isabelle Augenstein
\\
\bigskip
University of Copenhagen, Department of Computer Science, Copenhagen, Denmark
\\
\bigskip
*Corresponding author: dw@di.ku.dk

\end{flushleft}
\section*{Abstract}
Selecting an effective training signal for machine learning tasks is difficult: expert annotations are expensive, and crowd-sourced annotations may not be reliable. Recent work has demonstrated that learning from a distribution over labels acquired from crowd annotations can be effective both for performance and uncertainty estimation. However, this has mainly been studied using a limited set of soft-labeling methods in an in-domain setting. Additionally, no one method has been shown to consistently perform well across tasks, making it difficult to know a priori which to choose. To fill these gaps, this paper provides the first large-scale empirical study on learning from crowd labels in the out-of-domain setting, systematically analyzing 8 soft-labeling methods on 4 language and vision tasks. Additionally, we propose to aggregate soft-labels via a simple average in order to achieve consistent performance across tasks. We demonstrate that this yields classifiers with improved predictive uncertainty estimation in most settings while maintaining consistent raw performance compared to learning from individual soft-labeling methods or taking a majority vote of the annotations. We additionally highlight that in regimes with abundant or minimal training data, the selection of soft labeling method is less important, while for highly subjective labels and moderate amounts of training data, aggregation yields significant improvements in uncertainty estimation over individual methods. Code can be found at \url{https://github.com/copenlu/aggregating-crowd-annotations-ood}

\section*{Introduction}
One of the first concerns in supervised machine learning is how to define, collect, and use labels as training data for a given task. There are many tradeoffs associated with this decision, including the cost, the number of labels to collect, the time to collect those labels, the source of those labels, the accuracy of those labels with respect to the task under consideration, and the quality of models that are trained on those labels with respect to performance on e.g. estimating predictive uncertainty. These tradeoffs arise based on decisions about how the labels are collected (e.g. crowdsourcing, expert labeling, distant supervision) and how they are trained on in practice, for example as one-hot categorical labels (hard labeling) or as a distribution (soft labeling).

A large body of literature examines all facets of this question~\cite{DBLP:journals/jair/UmaFHPPP21}, with much work suggesting that using soft-labels for classification tasks improves model accuracy and uncertainty estimation~\cite{DBLP:conf/iccv/PetersonBGR19,DBLP:conf/hcomp/UmaFHPPP20,DBLP:conf/naacl/FornaciariUPPHP21}. Here, models are trained to minimize a loss with respect to their predictive distribution and a distribution over labels obtained from crowd annotations~\cite{DBLP:conf/hcomp/UmaFHPPP20}. While this has been shown to improve model generalization to unseen domains for vision tasks in some cases~\cite{DBLP:conf/iccv/PetersonBGR19}, no work has systematically compared how different soft-labeling schemes affect performance under distribution shift, where proper uncertainty estimation is paramount for decision making. We seek to fill this gap, providing a broad study into soft-labeling techniques from crowd-annotation in the out-of-domain setting across 8 methods, 4 vision and language tasks, and 7 datasets. Here, out-of-domain includes both data sourced from different corpora for the same high level task, and annotations sourced from different populations.

Such a comparison has been performed using a set of 4 soft-labeling methods in the \textit{in-domain} test setting in previous work~\cite{DBLP:conf/naacl/FornaciariUPPHP21, DBLP:journals/jair/UmaFHPPP21}, revealing that there is no clear best method across tasks. We find that this inconsistency also holds in the out-of-domain case for a larger set of soft-labeling approaches, as differences between methods are either insignificant or significantly better/worse depending on the task.
Given this, we explore whether \textit{aggregating} soft-labels, i.e. combining the distributions acquired from different methods into a single distribution, can produce better classifiers on out-of-domain data than using any single distribution. In doing so, we find that training classifiers on aggregated soft labels can generally achieve improvements in (predictive) uncertainty estimation while maintaining good raw performance. More specifically, we find that uncertainty estimation is significantly improved by aggregation when datasets have highly subjective labels, and in some cases when there is a moderate amount of training data. Otherwise, aggregation can help to overcome the inconsistency of individual soft labeling approaches.

In sum, we make the following contributions:%
\begin{enumerate}[noitemsep, label=\arabic*)]
    \item A comparison of soft-labeling techniques for learning from crowd annotations for 4 vision and language tasks across 7 datasets in the out-of-domain test setting, including text classification (recognizing textual entailment and toxicity detection), sequence tagging (part-of-speech tagging), and image classification;
    \item A simple intervention, namely averaging, which offers more consistent performance across tasks and significantly better uncertainty estimates in some settings;
    \item Insights into uncertainty estimation, performance, and calibration of models trained on soft labels in multiple settings.
\end{enumerate}

\section*{Related Work}

\paragraph{Learning from Crowd-Sourced Labels}
An efficient way to collect training data for a new task is to ask crowd annotators on platforms such as Amazon Mechanical Turk to manually annotate training data. How to select an appropriate training signal from these noisy crowd labels has a rich set of literature (e.g. see the survey from \cite{DBLP:journals/tacl/PaunCCHKP18}). Many of these studies focus on Bayesian methods to learn a latent distribution over the true class for each sample, influenced by factors such as annotator behavior~\cite{DBLP:conf/naacl/HovyBVH13,dawid1979maximum} and item difficulty~\cite{DBLP:conf/nips/WhitehillRWBM09}, and selecting the mean of this distribution as the final label. However, selecting a single true label discards potentially useful information regarding the uncertainty over classes inherent in many tasks, for example where items can be especially difficult or ambiguous~\cite{DBLP:conf/chi/GordonZPHB21}. Recent work has looked into how to learn directly from crowd-annotations~\cite{DBLP:journals/jair/UmaFHPPP21,DBLP:conf/uai/SucholutskyBCMP23}. The work of \cite{DBLP:conf/iccv/PetersonBGR19} demonstrated that learning directly from crowd annotations treated as soft-labels using the softmax function leads to better out of distribution performance in computer vision. This line of work has been followed by~\cite{DBLP:conf/hcomp/UmaFHPPP20} and \cite{DBLP:conf/naacl/FornaciariUPPHP21} in NLP, looking at the use of the KL divergence as an effective loss. The survey of \cite{DBLP:journals/jair/UmaFHPPP21} provides an extensive set of experiments comparing methods for learning from crowd labels. What has not been done is a systematic comparison of different soft-labeling methods in the out of domain setting.
We fill this gap in this work, and propose aggregation as a way to acquire more consistent and robust performance than previous methods without requiring new annotations or learning methods. Additionally, the work of \cite{DBLP:conf/uai/SucholutskyBCMP23} seeks to understand the impact of soft labels on representation learning, including out-of-distribution generalization, by eliciting different types of hard and soft judgements directly from crowd workers. Similar to \cite{DBLP:conf/iccv/PetersonBGR19}, they demonstrate that soft-labels can benefit out-of-distribution performance for image classification tasks. Our work adds to this by focusing on understanding out-of-distribution performance using different algorithmically acquired soft-labels from hard crowd-annotations across a diverse set of tasks and modalities.%

\paragraph{Learning from Soft Labels} Multiple lines of work investigate learning from soft-labels in general, with the closest lines of work to ours being knowledge distillation and label smoothing. Knowledge distillation seeks to build compact but robust models by training them on the probability distribution learned by a much larger teacher network~\cite{DBLP:conf/nips/BaC14,DBLP:journals/corr/HintonVD15}. The goal is to impart the ``dark knowledge'' contained in the distribution learned by the larger network, which can indicate similarities between features and classes if the output from the classifier is well calibrated (e.g. via temperature scaling~\cite{DBLP:journals/corr/HintonVD15} or ensembling~\cite{DBLP:journals/corr/HintonVD15,DBLP:journals/corr/abs-2012-09816}). 
Label smoothing~\cite{DBLP:conf/cvpr/SzegedyVISW16} seeks to regularize model training by inducing soft labels from hard labels and training on these directly. Generally this is accomplished through a weighted average of the one-hot distribution induced by gold labels and a purely uniform distribution, and yields classifiers with improved uncertainty estimation and calibration~\cite{muller2019does}. Inspired by this, we compare several methods to obtain soft labels from crowd annotations and investigate the effect of combining these labels with respect to out-of-domain generalization. 

\section*{Methods}
We build on a rich literature on learning from crowd annotations, particularly from soft-labels: distributions over classes obtained from annotations as opposed to selecting a single categorical label. In this, samples have their probability mass distributed over multiple classes, which can help regularize training. 
We present several well-studied methods for learning from crowd-labels, adding to this literature by analyzing their performance when considering generalization to out of domain data. Then, we propose to further improve the quality of labels by averaging across their distributions, which we will demonstrate leads to classifiers with improved predictive uncertainty estimation.

\subsection*{Soft Labeling Methods}
\label{sec:soft_labels}
We experiment with seven widely used methods for obtaining soft labels, including a mix of deterministic and Bayesian approaches. Each method makes different prior assumptions on the distribution of labels, thus reflecting different explanations of label uncertainty~\cite{DBLP:journals/tacl/PaunCCHKP18}. We present a brief overview of each method in the following. We use the implementation of these methods in \cite{CrowdKit} for our experiments.

\paragraph{Standard Normalization} The standard normalization scheme presented in~\cite{DBLP:conf/hcomp/UmaFHPPP20} obtains soft-labels for a given sample by transforming a set of crowd-sourced labels directly into a probability distribution. This is done by normalizing the number of votes given to each label by the total number of annotations for a given sample, as described in \autoref{eq:standard_norm}.
\begin{equation}
\label{eq:standard_norm}
    p_{stand}(i,c) = \frac{C_{i,c}}{\sum_{\hat{c}}C_{i,\hat{c}}}
\end{equation}
where $C_{i,c}$ is the number of votes label $c$ received for item $i$.

\paragraph{Softmax Normalization} The standard normalization scheme does not distribute probability mass to any label which receives no votes from any annotator. The works of~\cite{DBLP:conf/iccv/PetersonBGR19} and \cite{DBLP:conf/naacl/FornaciariUPPHP21} propose to use the softmax function directly from label vote counts as a way to obtain soft labels for a given sample, as in \autoref{eq:softmax_norm}. 
\begin{equation}
\label{eq:softmax_norm}
    p_{soft}(i,c) = \frac{e^{C_{i,c}}}{\sum_{\hat{c}}e^{C_{i,\hat{c}}}}
\end{equation}

\paragraph{Worker Agreement with Aggregate (Wawa)} 
Wawa~\cite{CrowdKit} is a deterministic method which performs a weighted averaging of the annotations based on estimated worker skills. A worker's skill is calculated as their average agreement with the majority vote on any given label. In this, the distribution is obtained as:

\begin{equation*}
    w_{a} = \frac{1}{M}\sum_{m}\mathbb{I}[a_{m} = \hat{a}_{m}]
\end{equation*}
\begin{equation}
    p_{wawa}(i,y) = \frac{1}{N}\sum_{a}w_{a}*y_{i,a}
\end{equation}
where $a$ is a given annotator, $a_{m}$ is the label that annotator $a$ assigns to item $m$, $\hat{a}_{m}$ is the majority vote for item $m$, $y_{i,a}$ is 1 if annotator $a$ assigned label $y$ to item $i$ (otherwise 0), and $\mathbb{I}[\cdot]$ is the indicator function.

\paragraph{ZeroBasedSkill (ZBS)} 
ZBS~\cite{CrowdKit} performs a similar operation to Wawa, but instead learns a worker's skill parametrically by minimizing the mean-squared error between the worker's accuracy with respect to the majority vote and their estimated skill. In other words,
\begin{equation*}
    \mathbf{w} = \min_{\mathbf{w}}\frac{1}{2}\sqrt{\sum_{a}\sum_{i}(w_{a} - \frac{1}{N}\sum_{j}w_{j}*y_{i,j}})^{2}
\end{equation*}
\begin{equation}
    p_{ZBS}(i,y) = \frac{1}{N}\sum_{a}w_{a}*y_{i,a}
\end{equation}

\paragraph{Dawid \& Skene (DS)} A common method for aggregating crowd-sourced labels into a single ground-truth label is to treat the true label as a latent variable to be learned from annotations. Several models have been proposed in the literature to accomplish this~\cite{dawid1979maximum,DBLP:conf/naacl/HovyBVH13,DBLP:conf/nips/WhitehillRWBM09}, often accounting for different aspects of the annotation problem such as annotator competence and item difficulty. One such method is the Dawid and Skene model~\cite{dawid1979maximum}, a highly popular method across fields for acquiring labels from crowd-annotations, which focuses in particular on modeling the true class based on each annotator's ability to correctly identify true instances of a given class. 
In other words, the model is designed to explain away inconsistencies of individual annotators. The model is given as follows:

\begin{equation}
\begin{split}
    z_{i} &\sim \text{Cat}(\pi_{i}) \\
        p(y_{i,a} = k | z_{i} = c) &= \mathbf{E}^{a}[k, c]
\end{split}
\end{equation}
where $\pi_{i}$ is a parameter indicating the probability distribution for item $i$, $\mathbf{E}^{a}$ is a $K\times K$ parameter matrix for annotator $a$, and $\mathbf{E}^{a}[k, c]$ is the probability that worker $a$ annotates label $k$ when the true label is $c$. To obtain a soft label for a given sample $i$ from this model, we simply use the posterior distribution of the latent variable $z_{i}$ i.e. $p_{DS}(i,c) = p(z_{i} = c)$.

\paragraph{Generative model of Labels, Abilities, and Difficulties (GLAD)} GLAD~\cite{DBLP:conf/nips/WhitehillRWBM09} extends the model from \cite{dawid1979maximum} by including item difficulty as a latent variable. As such, with observed annotations $y_{i,a}$, latent classes $z_{i}$, latent (inverse) item difficulties parameterized by $\beta_{i}$, and latent annotator competence variables parameterized by $\alpha_{a}$, the full model is given as:
\begin{equation}
    \begin{split}
        z_{i} &\sim \text{Cat}(\pi_{i}) \\
        p(y_{i,a} = k | z_{i} = c) &= \begin{cases}
                f(i,a) & k = c\\
                \frac{1 - f(i,a)}{K - 1} & k \neq c
        \end{cases}\\
        f(i,a) &= \frac{1}{1 + \exp(-\alpha_{a}\beta_{i})}
    \end{split}
\end{equation}
Similarly to \cite{dawid1979maximum}, we use the posterior distribution of the latent variable $z_{i}$ to acquire the soft label i.e. $p_{GLAD}(i,c) = p(z_{i} = c)$.

\paragraph{MACE} Finally, Multi-Annotator Competence Estimation (MACE, \cite{DBLP:conf/naacl/HovyBVH13}) is another Bayesian method popular in NLP which focuses specifically on explaining away poor performing annotators. It does this by learning to differentiate between annotators that are likely follow the global labeling strategy of selecting the true underlying label, from those which follow a labeling strategy which deviates from this e.g. spamming a single label for every example. To do this, it learns a distribution over the true label for each sample, as well as the likelihood that each annotator is faithfully labeling each sample. The likelihood for this model is given as follows.
\begin{equation}
    \begin{split}
        z_{i} &\sim \text{Cat}(\pi_{i}) \\
        T_{a} &\sim \text{Cat}(\epsilon_{a}) \\
        S_{i,a} &\sim \text{Bern}(\theta_{a}) \\
        p(y_{i,a} = c | z_{i}, T_{a}) &= \theta_{a}p(z_{i} = c) + (1 - \theta_{a})p(T_{a} = c)
    \end{split}
\end{equation}
Where $\theta_{a}$ is the parameter of a Bernoulli distribution indicating if annotator $a$ is spamming or not, $\epsilon_{a}$ are the parameters of a Categorical distribution encoding the local labeling strategy of annotator $a$, and $\pi_{i}$ are the parameters of a Categorical distribution representing the true label of item $i$. We use $p(z_{i} = c) = p_{MACE}(i,c)$ as the soft label for each item $i$. Additionally, each Bayesian model (DS, GLAD, and MACE) is learned using expectation maximization (EM).

\subsection*{Aggregating Soft Labels}
Training on soft labels as opposed to hard labels is well-motivated in the literature for improving out-of-domain performance. As an example, label smoothing~\cite{DBLP:conf/cvpr/SzegedyVISW16} has been shown to improve model calibration and generalization while making relatively uninformative assumptions about the distribution over labels~\cite{muller2019does}. In the label smoothing setup, training labels are generated as a weighted average between a one-hot distribution using the gold label and a uniform distribution over labels:
\begin{equation*}
    p_{ls}(i,c) = (1 - \epsilon)\mathbf{\mathbb{I}}_{K}(y_{i} = c) + \frac{\epsilon}{K}\mathbf{\mathbb{I}}_{K}(y_{i} \neq c)
\end{equation*}
where $\mathbf{\mathbb{I}}_{K}$ is the indicator function.

As opposed to label smoothing, each of the methods we have presented estimates a posterior distribution over labels using priors informed by various aspects of the crowd annotation problem. However, the prior assumptions made by each of these models may induce soft labels which are more or less appropriate given the task and data. In order to produce a set of soft labels which capture the uncertainty of these different posteriors, we propose to aggregate these distributions into a mixture distribution i.e. through averaging:
\begin{equation}
    p_{agg}(i,c) = \frac{1}{|\mathcal{D}|}\sum_{p_{d}\in\mathcal{D}}p_{d}(i,c)
\end{equation}
where $\mathcal{D}$ is a set containing some subset of the distributions outlined Equations 1-7. This can be seen as performing label smoothing over the soft labels using uniform weight $\epsilon = \frac{1}{|\mathcal{D}|}$ on the set of data-driven posterior distributions $\mathcal{D}$, contrasting with pure label smoothing which uses only one-hot categorical labels smoothed with a uniform distribution. Our hypothesis is that this will further improve model generalization and uncertainty estimation.

\section*{Experimental Setup}
\label{sec:experimental_setup}
Our experiments focus on the out-of-domain setting. %
We use pairs of datasets which capture the same high-level tasks and where the training data has only crowd-annotations available while the testing data has gold annotations. We use dataset pairs with one of two sources of domain shift: 1) input data sourced from different corpora; 2) labels acquired from different populations.

For all text tasks we use RoBERTa as our base network~\cite{DBLP:journals/corr/abs-1907-11692}
and for image classification we use a vision transformer (ViT)~\cite{wu2020visual} pretrained on Imagenet.%
This allows us to observe how the same soft-labeling techniques on the same network perform on different tasks. For the soft-labeling experiments we only use soft labels obtained using one of the crowd-labeling methods described in the Methods section and train using the KL divergence as the loss (as in previous work~\cite{DBLP:journals/jair/UmaFHPPP21,DBLP:conf/naacl/FornaciariUPPHP21}). For aggregation, we use all of the individual soft-labeling methods. We compare this with using a hard majority vote label trained with cross entropy loss, which is the common method of using crowd sourced annotations.
The tasks and datasets used in our experiments are described in the following paragraphs (full descriptions in the Supporting information).

\paragraph{Recognizing Textual Entailment (RTE)} is the first task we consider. In RTE, a model must predict whether a hypothesis is entailed (i.e. supported) by a given premise. For training, we use the Pascal RTE-1 dataset~\cite{DBLP:conf/mlcw/DaganGM05} with crowd-sourced labels from~\cite{DBLP:conf/emnlp/SnowOJN08} and for testing, we use the Stanford Natural Langauge Inference dataset (SNLI, \cite{DBLP:conf/emnlp/BowmanAPM15}).

\paragraph{Part-of-Speech Tagging (POS)} is a sequence tagging task to predict the correct POS for each token in a sentence. For training data, we use the Gimpel dataset from~\cite{DBLP:conf/acl/GimpelSODMEHYFS11} with the crowd-sourced labels provided by~\cite{DBLP:conf/acl/HovyPS14, DBLP:conf/acl/PlankHS14}. We use the publicly available sample of the Penn Treebank POS dataset~\cite{DBLP:journals/coling/MarcusSM94} accessed from NLTK~\cite{DBLP:conf/acl/Bird06} as our out-of-domain test set, which consists of 3,914 sentences from Wall Street Journal articles (100,676 tokens). 

\paragraph{Toxicity Detection} To test performance on a highly subjective task, we use the \href{https://www.kaggle.com/competitions/jigsaw-unintended-bias-in-toxicity-classification}{toxicity detection dataset} created as a part of the Google Jigsaw unintended bias in toxicity classification competition. %
We use the variant by~\cite{DBLP:journals/corr/abs-2205-00501}, who annotated 25,500 comments from the original Civil Comments dataset. The annotators are specifically selected and split into multiple rating pools based on self-indicated identity group membership (African American, LGBTQ). We randomly split the input data into training and test, and use the annotations from the original, unrestricted pool of annotators as gold labels to ensure a completely separate annotator pool that is not selected based on identity groups.

\input{tables/method_accuracy}

\input{tables/method_calibration}

\paragraph{Image Classification} Image classification is the task of categorizing an image into one of a set of discrete classes. As in \cite{DBLP:conf/iccv/PetersonBGR19}, for crowd-sourced data we leverage the CIFAR10H dataset and as an out-of-domain test set we use the CINIC10 dataset~\cite{DBLP:journals/corr/abs-1810-03505}. CIFAR10H consists of the 10,000 images from CIFAR10 test set which are reannotated by crowd-workers. CINIC10 consists of 210,000 images from the Imagenet dataset~\cite{DBLP:conf/cvpr/DengDSLL009} rescaled to the size of CIFAR10 images (32x32).

\section*{Results}
Our experiments examine the following research questions:
\begin{itemize}[noitemsep]
    \item \textbf{RQ1} How well do soft labels reflect gold labels?
    \item \textbf{RQ2}: Which soft-labeling methods are best in out-of-domain settings?
    \item \textbf{RQ3}: What is the effect of aggregating soft labels?
    \item \textbf{RQ4}: How does learning from soft-labels affect model calibration?
\end{itemize}

\subsection*{RQ1: How well do soft labels reflect gold labels?}

First, to understand the quality of labels produced by each soft-labeling method, we present the accuracy of the most probable class for each label compared to gold labels in \autoref{tab:dataset_accuracy} and uncertainty estimates in terms of negative log-likelihood in \autoref{tab:dataset_calibration}. Gold labels come from experts for all tasks except for toxicity detection, where gold labels are the majority vote of the larger, unrestricted pool of annotators. Aggregating soft labels produces best or near-best accuracy across each dataset. This contrasts with individual methods which vary in their rank depending on the dataset e.g. the method of \cite{dawid1979maximum} which produces the most accurate labels on CIFAR10H but the least accurate labels on Jigsaw. Negative log-likelihood also varies greatly by task and method. Aggregated soft-labels have generally good NLL, with the best NLL on RTE and CIFAR10H and near-best NLL on POS tagging and Jigsaw.

\subsection*{RQ2: Which soft-labeling methods are best in out-of-domain settings?}

We next evaluate the performance of models trained on labels from each soft-labeling method across two primary metrics: F1 score and calibrated log-likelihood (CLL, \cite{DBLP:conf/iclr/AshukhaLMV20}), measuring raw performance and predictive uncertainty respectively
We use the CLL in order to obtain a fair comparison between methods, as it first performs temperature scaling on a held-out portion of the test set and measures the temperature-scaled negative log-likelihood on the rest of the test set, averaging results over 5 splits
(for formal definition, see Supporting information). %
We then look at model calibration using reliability diagrams (for POS, RTE, and image classification) and distribution over total-variation distance (TVD, for toxicity detection) which is more appropriate on subjective annotation tasks~\cite{baan2022stop}. We discuss general observations from our results, and based on this provide answers for our research questions.

\subsubsection*{Uncertainty Estimation}
We first discuss how each method performs in terms of uncertainty estimation, measured using CLL. Results can be seen in the second column of each task in \autoref{tab:results}.

\input{tables/metrics}

\begin{figure}[h]
     \centering
     \includegraphics[width=0.95\linewidth]{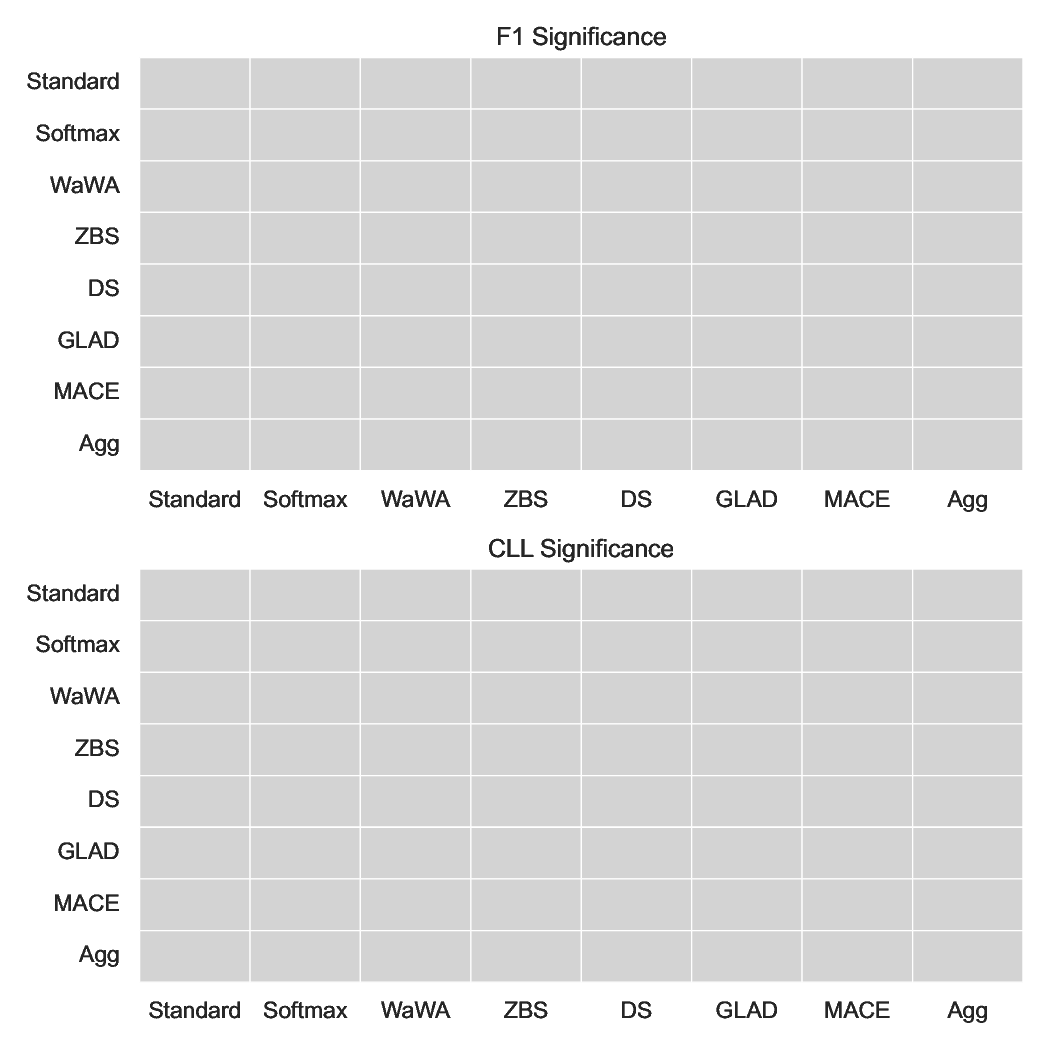}
     \caption{Significance testing for the RTE task. We apply the Bonferroni correction across the number of independent variables (N = 7; a more conservative estimate across the total tests N = 56 can be found in the supplemental information). Green indicates the method in the row is significantly better than the method in the column. Red indicates the method in the row is significantly worse than the method in the column. Grey indicates no statistically significant difference. }
     \label{fig:rte-sig}
\end{figure}

\begin{figure}[h]
     \centering
     \includegraphics[width=0.95\linewidth]{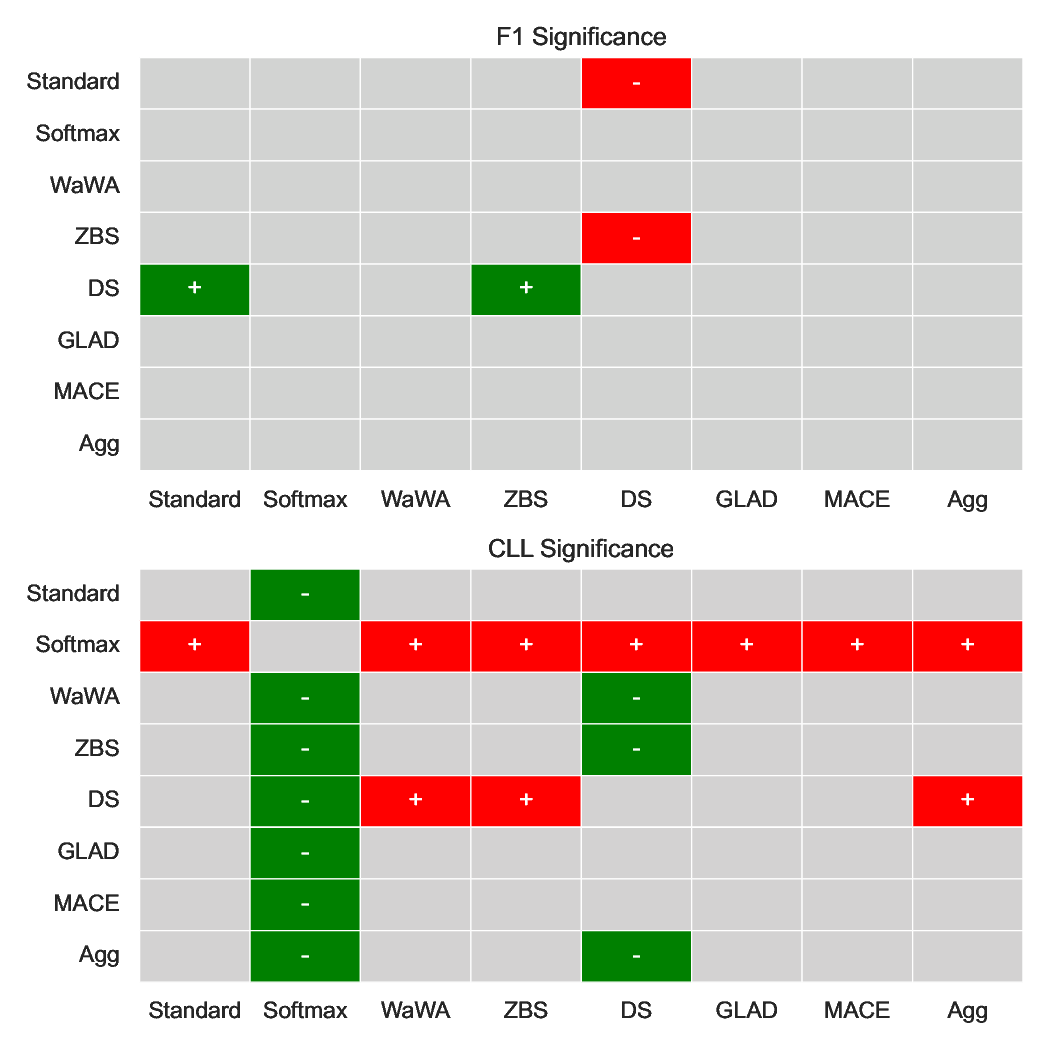}
     \caption{Significance testing for the POS task. We apply the Bonferroni correction across the number of independent variables (N = 7; a more conservative estimate across the total tests N = 56 can be found in the supplemental information). Green indicates the method in the row is significantly better than the method in the column. Red indicates the method in the row is significantly worse than the method in the column. Grey indicates no statistically significant difference. }
     \label{fig:pos-sig}
\end{figure}

\begin{figure}[h]
     \centering
     \includegraphics[width=0.95\linewidth]{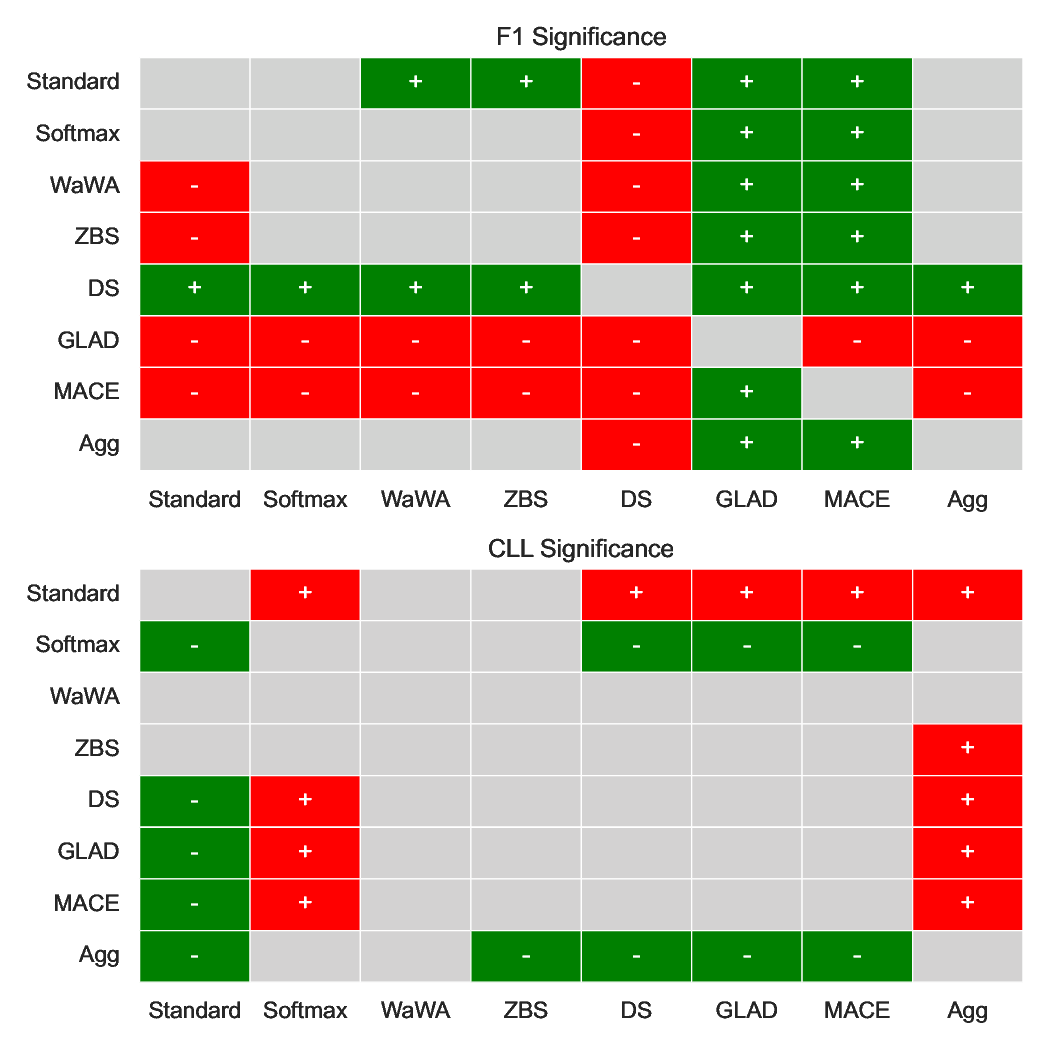}
     \caption{Significance testing for the Toxicity task. We apply the Bonferroni correction across the number of independent variables (N = 7; a more conservative estimate across the total tests N = 56 can be found in the supplemental information). Green indicates the method in the row is significantly better than the method in the column. Red indicates the method in the row is significantly worse than the method in the column. Grey indicates no statistically significant difference. }
     \label{fig:toxicity-sig}
\end{figure}

\begin{figure}[h]
     \centering
     \includegraphics[width=0.95\linewidth]{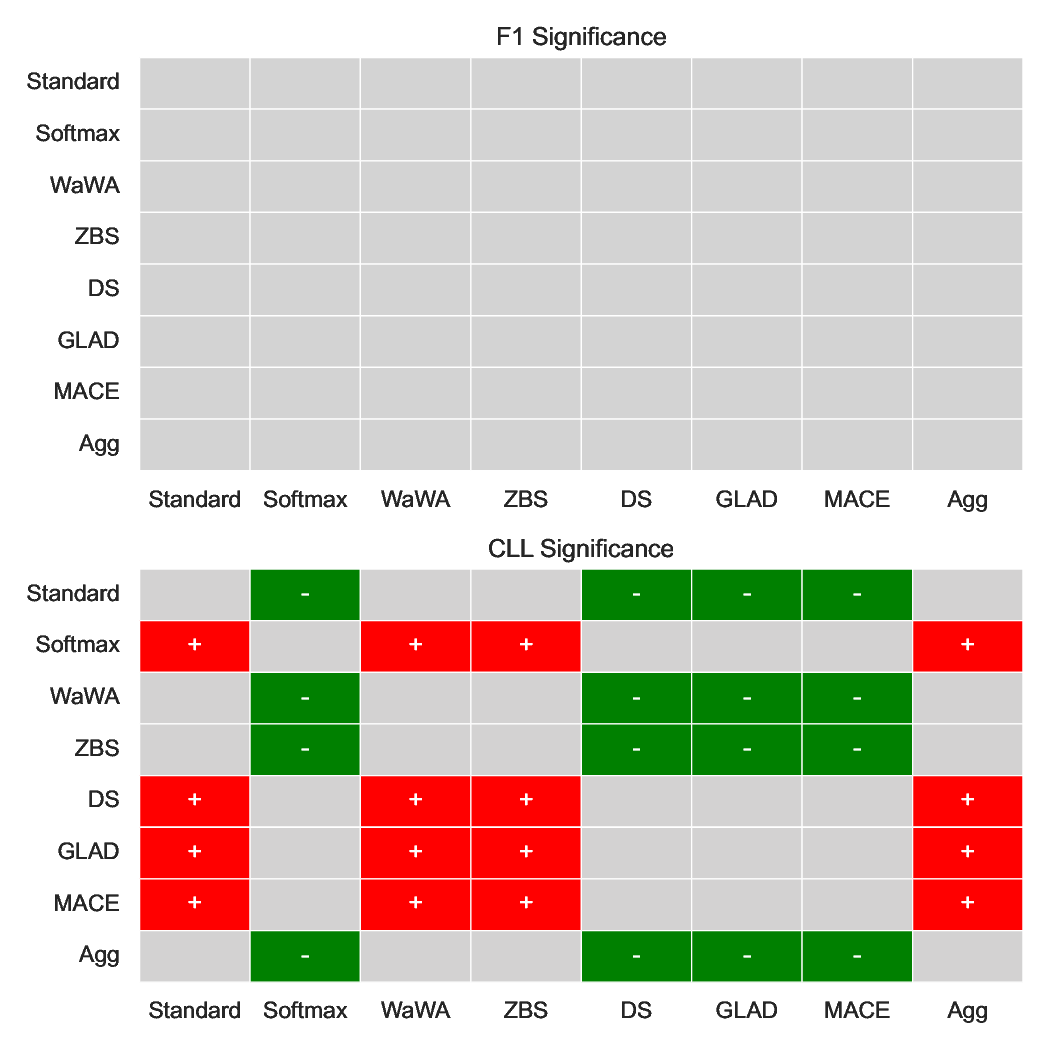}
     \caption{Significance testing for the Image Cls. task. We apply the Bonferroni correction across the number of independent variables (N = 7; a more conservative estimate across the total tests N = 56 can be found in the supplemental information). Green indicates the method in the row is significantly better than the method in the column. Red indicates the method in the row is significantly worse than the method in the column. Grey indicates no statistically significant difference. }
     \label{fig:img-sig}
\end{figure}

We find that aggregating soft labels leads to classifiers with the best uncertainty estimation on all text-based tasks and near-best performance on image classification. Additionally, aggregation is the only soft-labeling method which always provides better uncertainty estimates than using a majority vote hard-label. Individual soft-labeling methods will rank differently depending on the task e.g. using a softmax which has the second best performance on toxicity detection but the worst on POS tagging. The result of this is that individual soft-labeling methods generally yield better uncertainty estimates than a majority vote but this is not always guaranteed depending on the task and method. Aggregating the soft labels offers a way to improve upon this.

\subsubsection*{Raw Performance} 
Raw (macro) F1 scores are shown in the first column for each dataset (\autoref{tab:results}). First, we highlight the difficulty of the RTE task, which shows high variance in the results; this is due to the limited number of training samples (800), making it highly challenging to generalize to out-of-distribution data. Next, we see that using soft labels generally outperforms majority vote labels on each task. However, similarly with uncertainty estimation we find that each soft labeling method ranks differently depending on the task, and does not always outperform a majority vote. For example, using the standard distribution yields the best performance on RTE but the worst performance on POS tagging. Similarly, DS has the best performance on POS and toxicity detection but second worst on RTE. This highlights both the utility of using soft labels for training over majority vote, and the pitfalls of selecting an appropriate soft labeling method.

\subsubsection*{Discussion of Results}

\input{tables/significance_comparison}

\begin{figure}[h]
     \centering
     \includegraphics[width=0.95\linewidth]{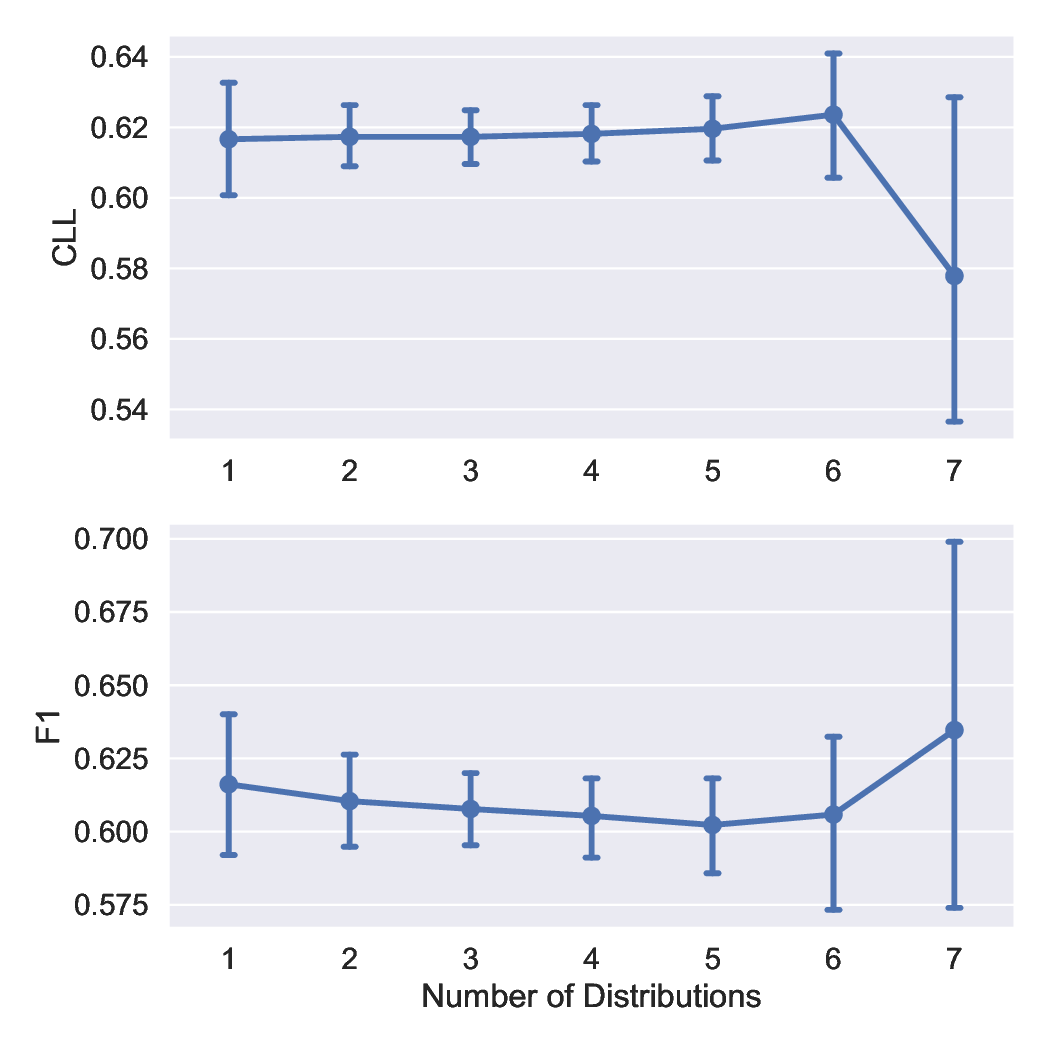}
     \caption{Comparison of the average CLL and F1 score on the RTE task using different combinations of distributions for aggreagation. Points are the average performance across all combinations of a given number of distributions and error bars are 95\% confidence intervals.}
     \label{fig:rte-subsets}
\end{figure}

\begin{figure}[h]
     \centering
     \includegraphics[width=0.95\linewidth]{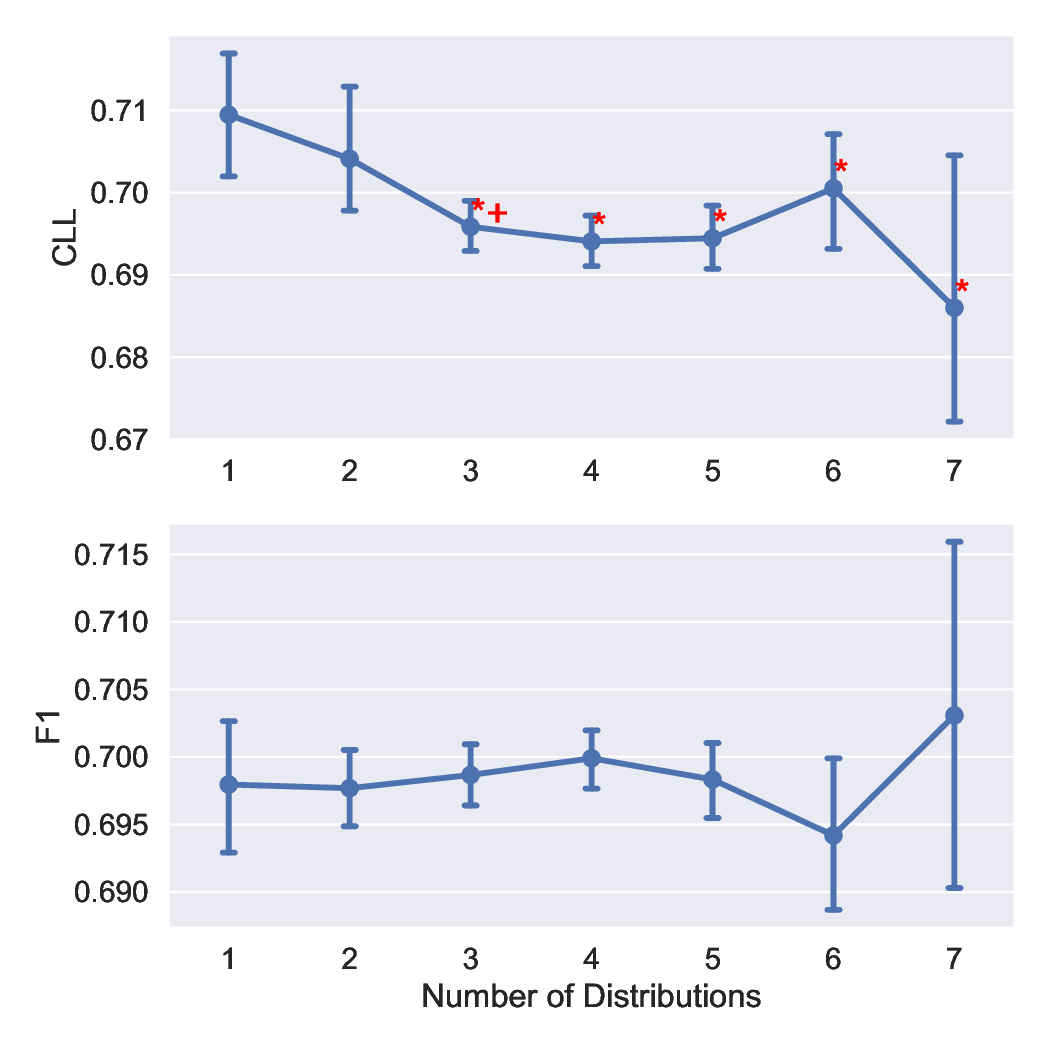}
     \caption{Comparison of the average CLL and F1 score on the POS tagging task using different combinations of distributions for aggreagation. Points are the average performance across all combinations of a given number of distributions and error bars are 95\% confidence intervals.}
     \label{fig:pos-subsets}
\end{figure}

\begin{figure}[h]
     \centering
     \includegraphics[width=0.95\linewidth]{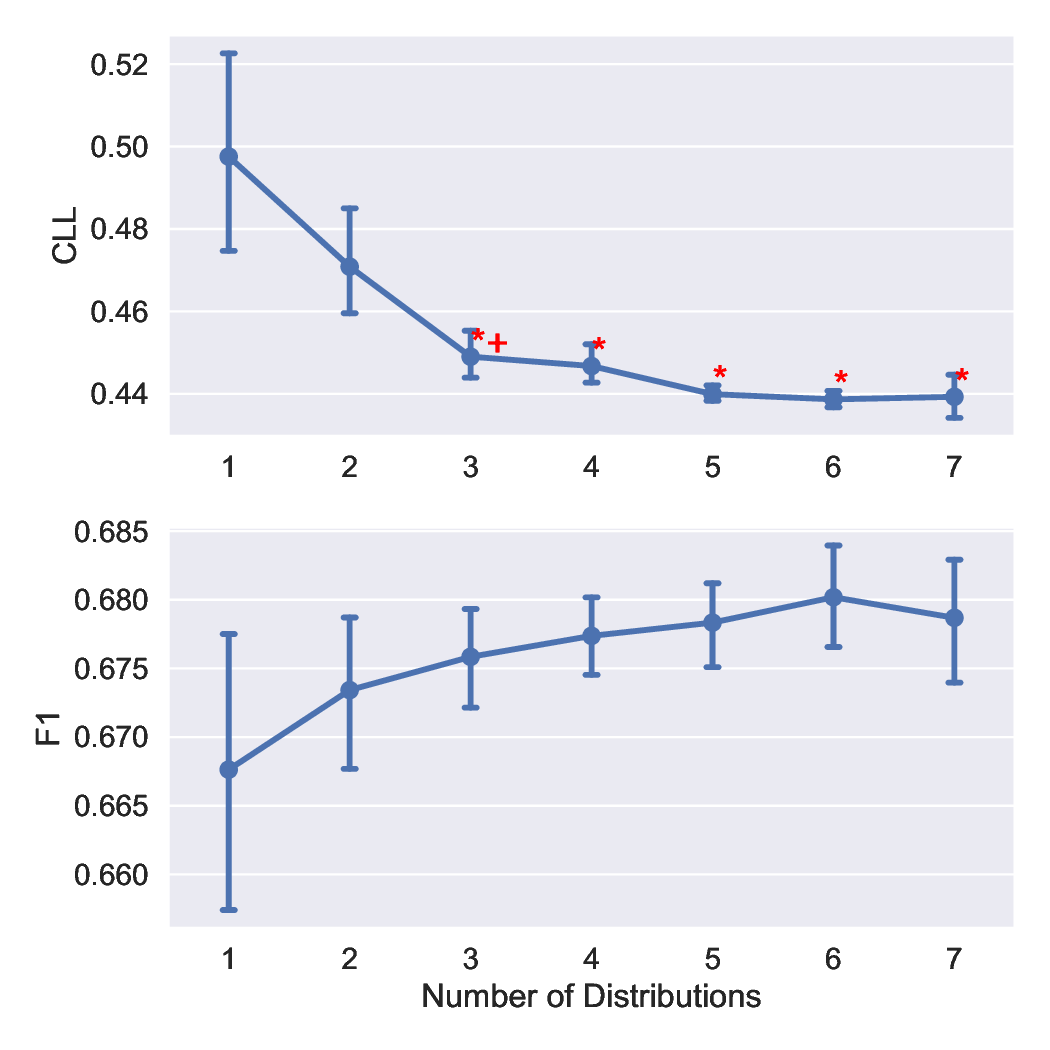}
     \caption{Comparison of the average CLL and F1 score on the Toxicity detection task using different combinations of distributions for aggreagation. Points are the average performance across all combinations of a given number of distributions and error bars are 95\% confidence intervals.}
     \label{fig:jigsaw-subsets}
\end{figure}

\begin{figure}[h]
     \centering
     \includegraphics[width=0.95\linewidth]{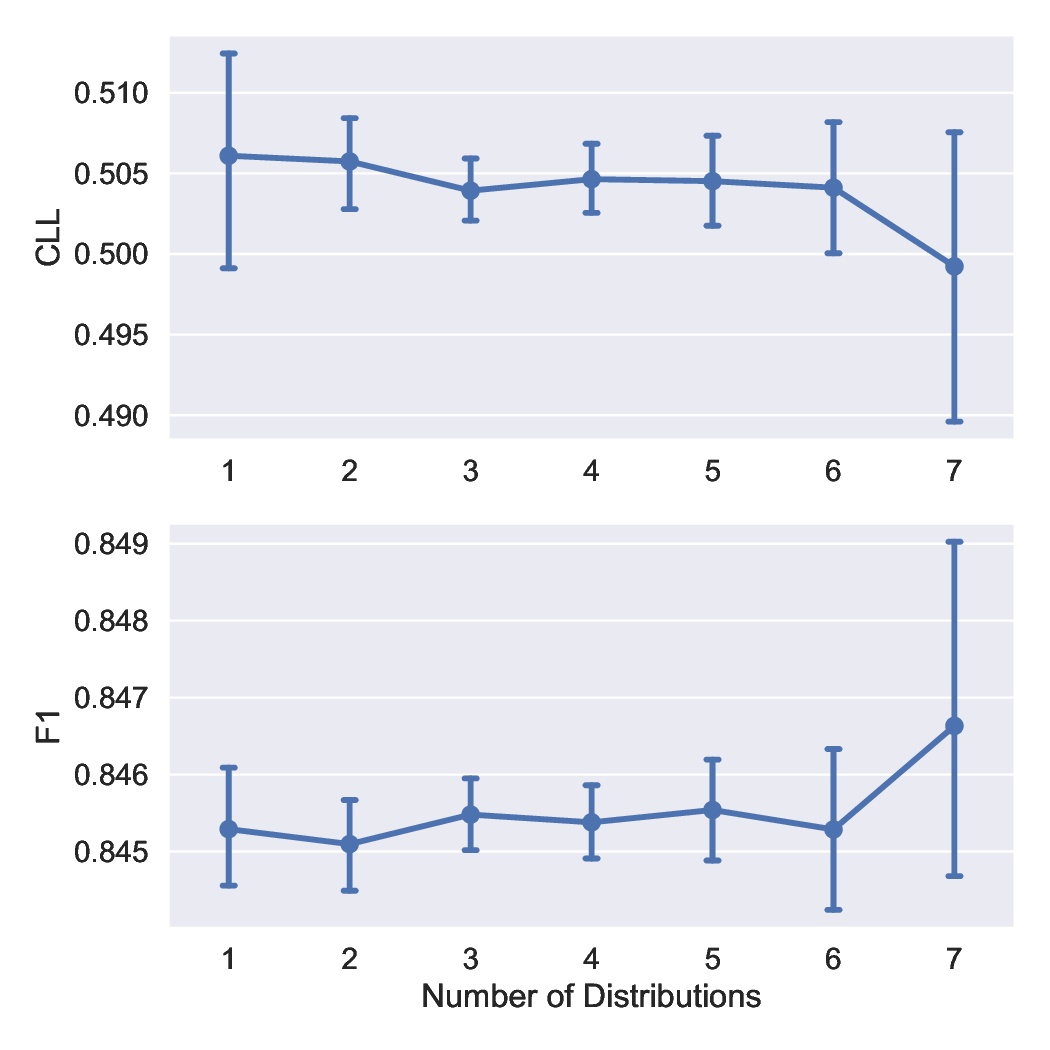}
     \caption{Comparison of the average CLL and F1 score on the image classification task using different combinations of distributions for aggreagation. Points are the average performance across all combinations of a given number of distributions and error bars are 95\% confidence intervals.}
     \label{fig:cifar10h-subsets}
\end{figure}

To contextualize our discussion, we measure if differences in performance are statistically significant at a level of p $<$ 0.05 using a Mann-Whitney U test~\cite{mann1947test}, and plot this for both F1 and CLL across all datasets in \autoref{fig:rte-sig}-\autoref{fig:img-sig}. We apply the Bonferroni correction across the number of independent variables (N = 7). A more conservative estimate across the total tests (N = 56) can be found in the supplemental information. For reference, the deterministic methods include: Standard, Softmax, Wawa, and ZBS. The Bayesian methods include DS, MACE, and GLAD. Majority voting does not result in soft-labels, so we consider this only as a baseline point of comparison.

For a high level perspective on both raw performance and uncertainty estimation, we first quantify each method in terms of how often it is statistically significantly better than other methods. We calculate this for all methods and separate F1 performance and uncertainty estimation into different scores in \autoref{tab:significance_comparison}. As we can see, DS is most often statistically significantly better than other methods in terms of raw performance, but is poorly calibrated as it is the second worst in terms of uncertainty estimation. Aggregating multiple soft distributions offers the best uncertainty estimates, while generally offering raw performance on par with the best methods. This is also seen in \autoref{fig:rte-sig}-\autoref{fig:img-sig}, where aggregation only has 1 instance of being worse in terms of F1 (on toxicity detection) and no instances of being worse in terms of uncertainty estimation (on image classification). Therefore, we find that aggregation offers the best tradeoff among all methods in terms of raw score and uncertainty estimation. 

We now zoom in to specific tasks and make more fine-grained recommendations. For RTE, we see that deterministic methods generally have greater mean F1 scores than Bayesian methods. Standard, Wawa, and ZBS outperform DS, MACE, and GLAD, with Standard as the best performing method, and Softmax outperforming at least DS and GLAD. For uncertainty estimation on RTE, the only trend we see is that aggregation may help. That being said, the limited amount of training data makes these differences not statistically significant. Therefore, in the low data regime, the choice of soft labeling approach will have limited impact, though a safe choice may be to go with the Standard method, which makes the least assumptions about the distribution underlying the annotations.

For POS tagging, the Bayesian approaches (DS, MACE, and GLAD) are on average better than all deterministic methods (Wawa, ZBS, Standard, and Softmax) in terms of F1. DS is statistically significantly better than Standard and ZBS. The only trend in uncertainty estimation is that aggregation leads to the best performance on average. Significance testing shows that Softmax is significantly worse than all other methods in terms of uncertainty estimation, and DS worse than WaWA and ZBS, showing that while DS leads to good raw performance, it also leads to a poorly calibrated output distribution. Therefore, the selection of soft labeling method here depends on whether good uncertainty estimation is a requirement; if not, then DS could be a good choice, otherwise aggregation or WaWA would be a good choice since they are not statistically significantly worse than DS in terms of raw performance while being significantly better in terms of uncertainty estimation.

For toxicity detection, which relies on subjective labels, we see that methods based on worker skill (MACE and GLAD for Bayesian approaches, and Wawa and ZBS for deterministic approaches) hurt performance. The drop (compared to Standard and Softmax) is less for Wawa and ZBS, but very large for MACE and GLAD compared to DS which makes no assumptions about worker skill or item difficulty. MACE and GLAD are statistically significantly worse than all other methods. The best performance is again DS, which is statistically significantly better than all other approaches, making it a strong choice here as well. For uncertainty estimation, aggregation leads to the best performance and is statistically significantly better than all other methods except for Softmax and WaWA. Therefore, both DS and aggregation are potentially appropriate for subjective tasks as they each offer a reasonable tradeoff between raw performance and uncertainty estimation, while approaches based on worker skill or item difficulty should be avoided.

For image classification, which has the largest set of training data, each method performs quite similarly, so (similarly to RTE) the choice of soft labeling method is not as critical. For raw performance, there is no statistically significant difference between any method. For uncertainty estimation, aggregation, WaWA, ZBS, and Standard are significantly better than all other methods, making them good choices.

\subsection*{RQ3: What is the Effect of Aggregating Soft Labels?}

We now look at the impact of the number of distributions used on model performance. We obtain soft labels for each combination of distribution and train models using six random seeds for all tasks in the study. We additionally compare to using individual soft labeling methods ($x = 1$ in each plot). Statistical significance is measured using a Mann-Whitney U test~\cite{mann1947test}. We mark two indicators of significance: a “*” indicates that aggregating this number of distributions is statistically significantly better than no aggregation (i.e., using only a single distribution). A “+” indicates statistical significance over the previous value marked with a “+” (or 1 for the first “+”). In other words, a “+” indicates progressive statistically significant improvement in performance. We again apply the Bonferroni correction with N = 7. Results on the impact of number of distributions are given in \autoref{fig:rte-subsets}-\autoref{fig:cifar10h-subsets}. 

\begin{figure*}[h]
     \centering
     \includegraphics[width=0.95\linewidth]{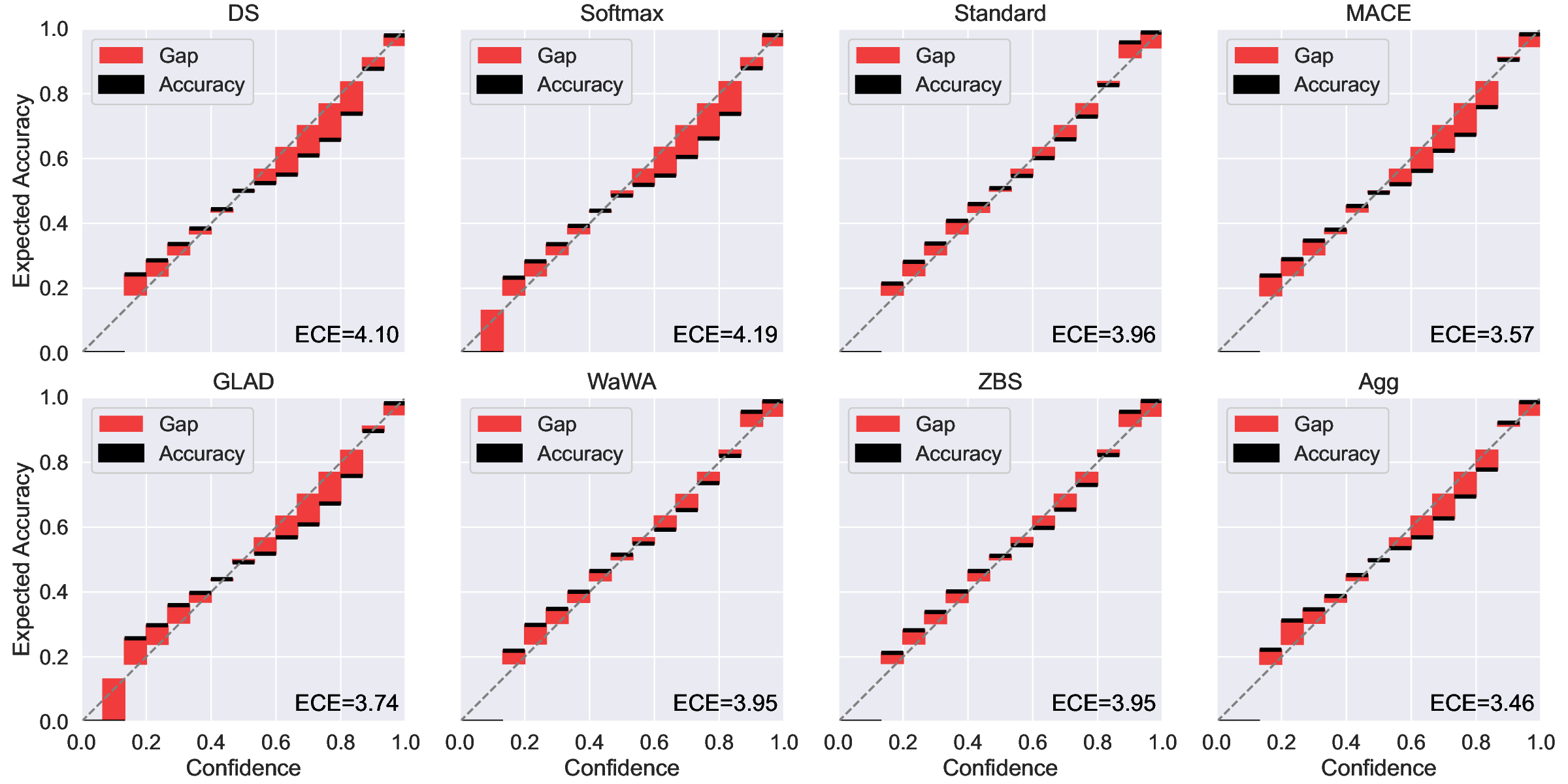}
     \caption{Reliability diagram and expected calibration error (ECE, displayed as \autoref{eq:ece} $\times$ 100) for each soft labeling method on image classification using the CINIC10 dataset. Black bars indicate the accuracy in the given bin and red bars indicate the gap between accuracy and confidence. Perfect calibration, where confidence and accuracy are equal, would follow the dotted line (i.e. a black bar over the line indicates under-confidence and a black bar under the line indicates over-confidence). We use the average of the logits produced by models trained with 10 different random seeds with no temperature scaling. Aggregating soft labels results in the best overall calibration.}
     \label{fig:img-reiability}
\end{figure*}
\begin{figure*}[h]
     \centering
     \includegraphics[width=0.95\linewidth]{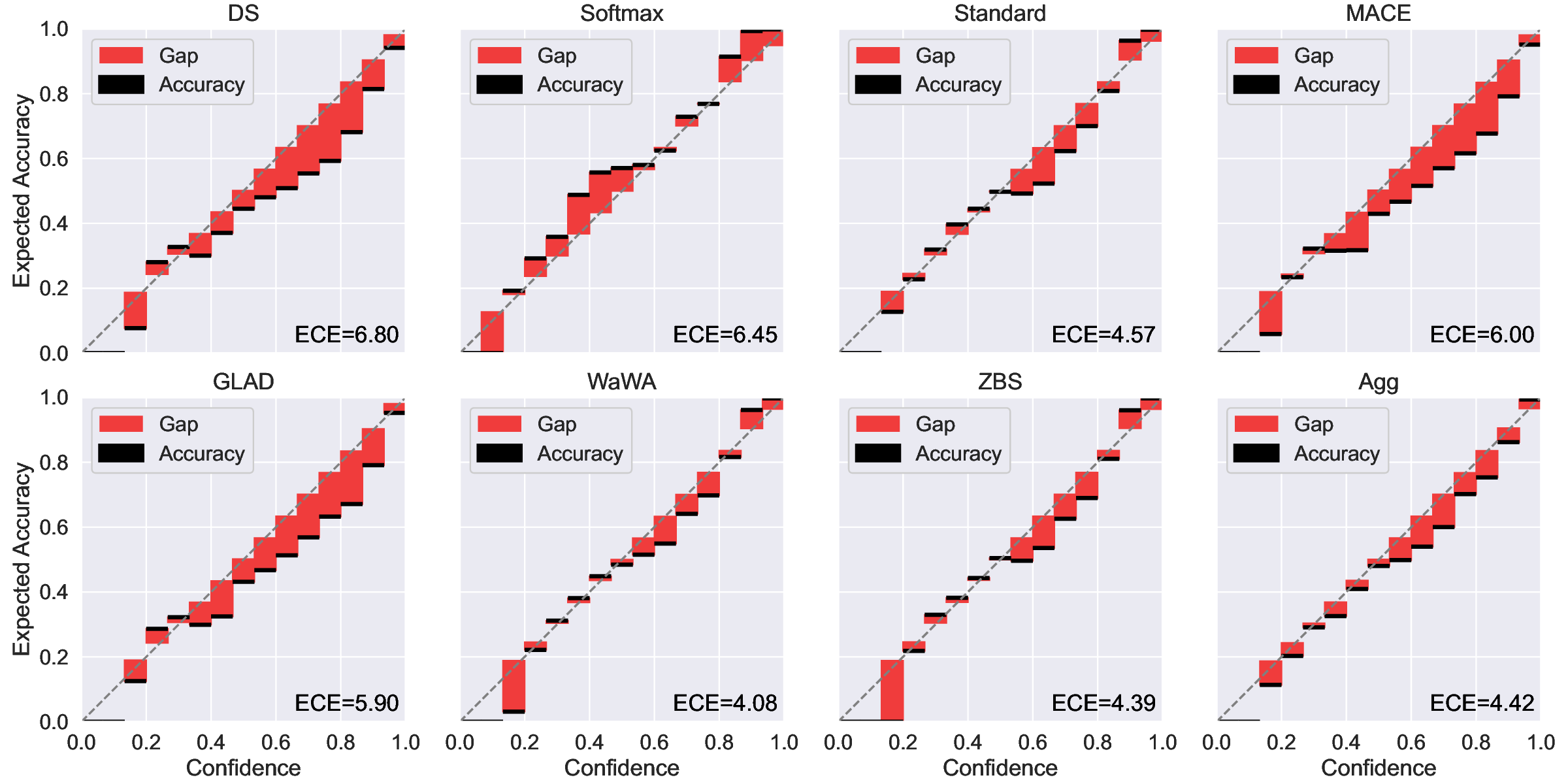}
     \caption{Reliability diagram and expected calibration error (ECE, displayed as \autoref{eq:ece} $\times$ 100) for each soft labeling method in POS tagging. Black bars indicate the accuracy in the given bin and red bars indicate the gap between accuracy and confidence. We use the average of the logits produced by models trained with 10 different random seeds with no temperature scaling. ECE for aggregation is comparable to the best performing methods (WaWA and ZBS). Models trained using aggregated soft labels have better calibration in both the low and high confidence regimes.}
     \label{fig:pos-reiability}
\end{figure*} 
\begin{figure*}[h]
     \centering
     \includegraphics[width=0.95\linewidth]{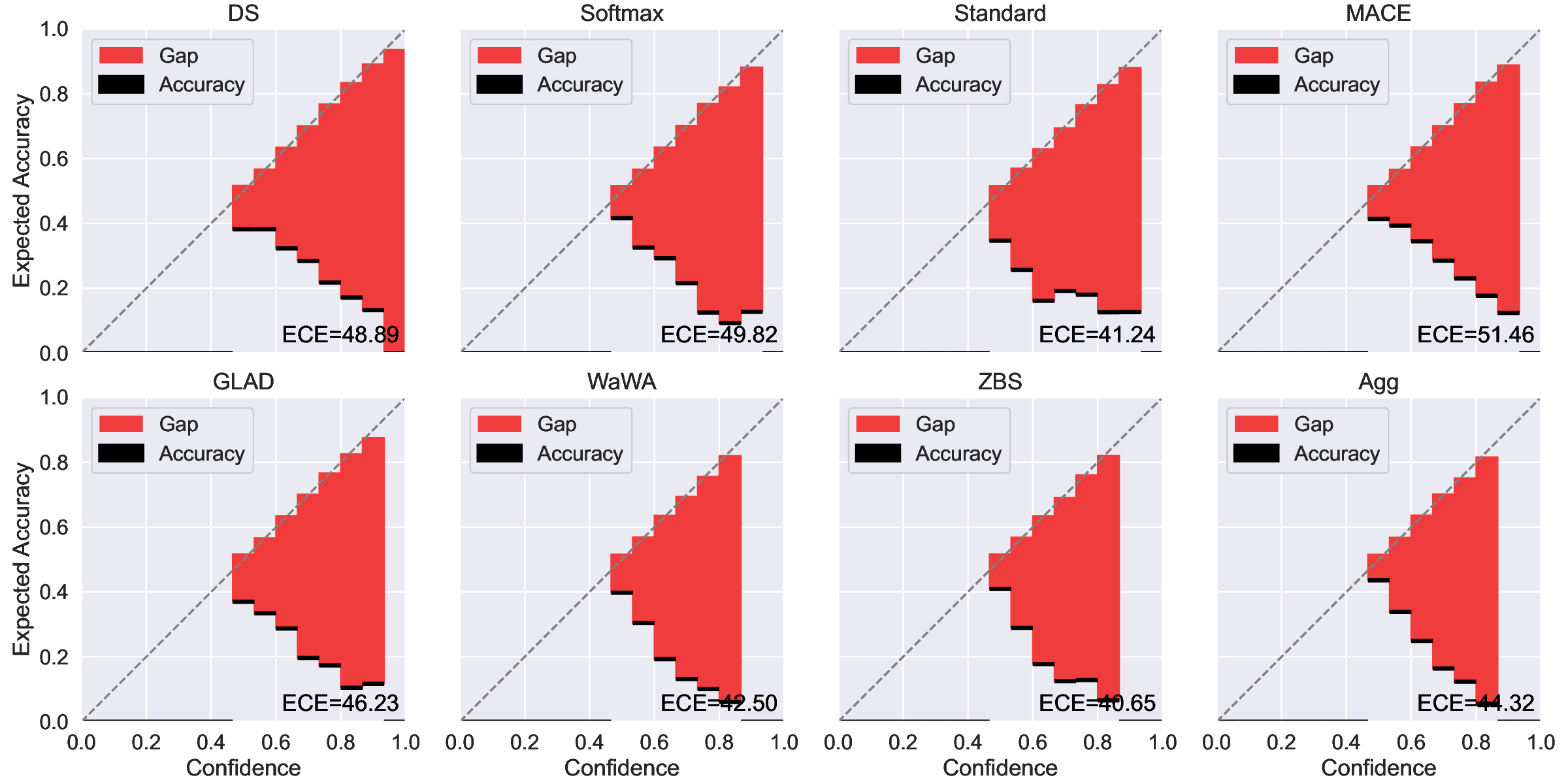}
     \caption{Reliability diagram and expected calibration error (ECE, displayed as \autoref{eq:ece} $\times$ 100) for each soft labeling method on RTE. Black bars indicate the accuracy in the given bin and red bars indicate the gap between accuracy and confidence. We use the average of the logits produced by models trained with 10 different random seeds with no temperature scaling. All models are highly overconfident, potentially as a result of the limited amount of training data with moderate reliability.}
     \label{fig:rte-reiability}
\end{figure*} 

We find that the combination of distributions can have a large impact on performance, and that adding distributions can improve performance. In the cases of extreme dataset size (both RTE and Image Cls.) we see no statistically significant difference between aggregating different number of distributions. For POS tagging, aggregating more than 3 distributions leads to statistically significant improvements over no aggregation for uncertainty estimation, with no statistical significance for F1 score. Finally, for toxicity detection we see statistically significant improvements for uncertainty, saturating at 5 distributions.

 Given this, in the medium data regime aggregating more distributions has clear benefits for uncertainty estimation, while for other tasks, the number of distributions aggregated has less impact. Additionally, we see that aggregation tends to perform better on average than using the individual soft labeling methods, as the performance of these methods can vary greatly for each task.

\subsection*{RQ4: How does learning from soft-labels affect model calibration?} 

Finally, we look into model calibration. To do so, we ensemble (average) the logits produced by each method across the 10 different seeds used in the previous experiments without temperature scaling, and use two different techniques of measuring calibration using these logits based on the task.

For POS, RTE, and image classification, we look at reliability diagrams and expected calibration error (ECE)~\cite{DBLP:conf/icml/GuoPSW17}. The reliability diagram shows the accuracy of each method for different levels of model confidence. This is done by first binning the model output probabilities based on the most confident class, and measuring the accuracy of the labels in each bin. A perfectly calibrated model will exhibit a 1:1 linear relationship between confidence and accuracy, with low confidence predictions having low accuracy and high confidence predictions having high accuracy. ECE formally measures this by taking a weighted average of the absolute difference between accuracy and confidence across bins (lower is better). Given $N$ bins where $b_{i}$ indicates the fraction of all test data in bin $i$, ECE is calculated as:
\begin{equation} 
\label{eq:ece}
    \sum_{i}^{N}b_{i}||\mathbf{p}_{i}-\mathbf{c}_{i}||_{1}
\end{equation}
where $\mathbf{p}_{i}$ is the accuracy of all predictions in bin $i$, $\mathbf{c}_{i}$ is the confidence of all predictions in bin $i$, and $\|\cdot\|_{1}$ is the L1-norm.
In our analysis, we use a bin size of 15 across methods and tasks.

The reliability diagrams and ECE for image classification is given in \autoref{fig:img-reiability}. For image classification we find that aggregating soft labels offers the best ECE compared to other methods. Additionally, each method yields a similar balance of under- and over-confident bins, mostly following a mostly linear trend. For POS tagging (\autoref{fig:pos-reiability}), most methods yield primarily over-confident classifiers with the exception of the softmax method which is primarily under-confident. Aggregation is most similar to the ZBS and Wawa methods, which offer better ECE and are less over-confident compared to other methods on this task. Additionally, aggregation provides better calibration in both the low- and high-confidence regimes compared to other methods. Finally, for RTE (\autoref{fig:rte-reiability}), all models tend to be overconfident potentially due to the limited amount of training data.

For toxicity detection we look at the distribution of total variation distance (TVD) between model predictions and the distribution over test labels. TVD is calculated as follows:
\begin{equation*}
    \text{TVD}(\mathbf{p}, \mathbf{q}) = \frac{1}{2}||\mathbf{p} - \mathbf{q}||_{1}
\end{equation*}
i.e. one half the L1 distance between the vector of probabilities of two categorical distributions $\mathbf{p}$ and $\mathbf{q}$. The motivation for using TVD as opposed to ECE is because toxicity detection is a highly subjective task, and as such ECE is an inappropriate metric due to its assumption of a single gold label~\cite{baan2022stop}. TVD instead measures a proper distance between the predicted probability distribution of a given model and the distribution induced by population-level human label variation. As such, we measure TVD between the probabilities output by each method 
 on each instance and the standard distribution over crowd annotations for that instance $p_{stand}(i,c)$ (\autoref{eq:standard_norm}). A model with better calibration to the human label variation will thus produce a distribution over TVD values more concentrated around 0, with perfect calibration being a Dirac delta at 0 (i.e., no deviation from human label variation).

\begin{figure*}[h]
     \centering
     \includegraphics[width=0.95\linewidth]{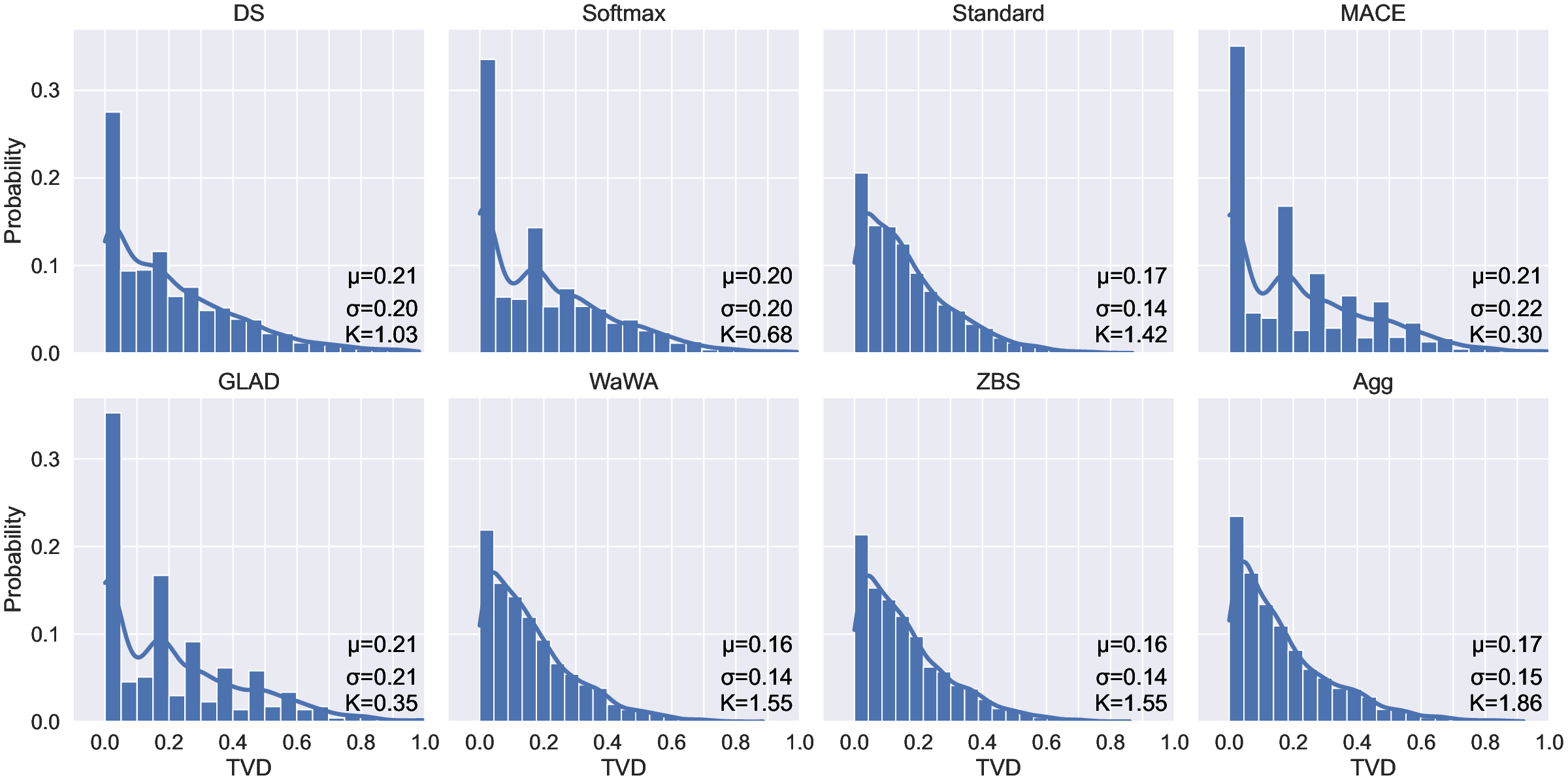}
     \caption{Distribution of total variation distance (TVD) between model predictions and original crowd-sourced annotations for the Jigsaw toxicity detection dataset. Perfect calibration is a TVD of 0, perfect miscalibration is a TVD of 1. ``K'' indicates the kurtosis of the distribution. Aggregation, standard, Wawa, and ZBS produce distributions with low mean and standard deviation compared to other methods.}
     \label{fig:jigsaw-tvd}
\end{figure*}

\autoref{fig:jigsaw-tvd} presents the distribution over TVD for the Jigsaw dataset using the original annotations coming from a broad, unrestricted set of human annotators. We find that most soft-labeling methods yield high peaks at 0 (indicating good calibration) but differ in terms of their tails. For example, while DS, MACE, and softmax have the highest peaks at 0, they result in more samples which are miscalibrated overall. Aggregation produces similarly calibrated classifiers to the best performing individual methods but with higher kurtosis, potentially due to the slightly higher peak at 0. 

\section*{Discussion and Conclusion}
In this work we provide a systematic comparison of soft-labeling techniques using crowd-sourced labels and demonstrate their utility on out-of-domain performance for several text and vision tasks. The out-of-domain setting allows us to observe how learning from crowd-sourced soft-labels enables generalization to unseen domains of data, where uncertainty estimation is critical for good decision making. Given than no consistent best performing individual method appears, we propose to aggregate these labels using a simple average. In doing so, we have answered the following research questions:

\textbf{RQ1: Do soft-labels reflect gold labels?} We demonstrate that aggregating soft labels produce best or near-best accuracy and negative log-likelihood compared to gold labels across all tasks.

\textbf{RQ2: Best individual methods for OOD performance.} Similarly to the in-domain case~\cite{DBLP:journals/jair/UmaFHPPP21}, we find that among individual soft-labeling techniques no consistent and clear best performer arises in the out-of-domain setting. These individual methods sometimes produce better classifiers than using a simple majority vote, but not always. 
\\\textbf{RQ3: Effect of aggregation.} Aggregating soft-labels leads to the most consistent performance across tasks in the out of distribution setting. It offers best or near-best uncertainty estimation and is the only method which always yields classifiers with better uncertainty estimates than a majority vote. Raw performance is generally consistent across tasks and better than using a majority vote. Adding more distributions generally leads to performance improvements, particularly for uncertainty estimation in challenging tasks (limited training data and higher subjectivity).
\\\textbf{RQ4: Model Calibration.} For image classification and POS tagging, models tend to be overconfident despite training on soft distributions. On toxicity detection, there are stark differences between methods in terms of how diffuse their distribution over TVD is. Similar to uncertainty estimation, aggregating soft labels yields best or near best calibration across these tasks, with the best ECE for image classification, and comparable ECE and TVD to the best methods for POS tagging, Toxicity detection, and RTE.

We conclude that aggregation constitutes a simple intervention to acquire reliable soft-labels from crowd-annotations which outperform majority vote labels alone on out-of-domain data. We hope that this work will spur further research on learning under human label variation, as this variation contains a rich pool of information that is useful for building robust classifiers that reflect real human ratings.

\section*{Limitations}
While we generally find that aggregation leads to better uncertainty estimation, we note the following limitations. First, we see that improvements in uncertainty estimates are not always statistically significant across all methods and datasets, though on the whole there is a tendency towards improvement. Second, the datasets we use in this study, as well as the out-of-distribution settings, are highly diverse. That being said, there are no highly consistent trends \textit{across} datasets for any individual methods, so more in-depth studies of specific types of domain shifts (e.g., changing annotator pools or changing document type) or specific dataset characteristics (e.g. few-shot datasets, or very large datasets) could help illuminate more discernible trends as to which types of soft-labeling methods are best for which scenarios. 


\section*{Supporting information}

\subsection*{Full Dataset Descriptions}
\label{sec:dataset-descriptions}
\paragraph{Recognizing Textual Entailment (RTE)} The first task we consider is recognizing textual entailment (RTE). In the RTE task, a model must predict whether a hypothesis is entailed (i.e. supported) by a given premise. For training, we use the Pascal RTE-1 dataset~\cite{DBLP:conf/mlcw/DaganGM05} with crowd-sourced labels from~\cite{DBLP:conf/emnlp/SnowOJN08}. The dataset consists of 800 premise-hypothesis pairs annotated by 164 different non-expert annotators with 10 annotations per pair. The inter-annotator agreement (IAA) is 0.629 (Fleiss $\kappa$). As an out-of-domain test set, we use the Stanford Natural Langauge Inference dataset (SNLI)~\cite{DBLP:conf/emnlp/BowmanAPM15}, where we transform the task into binary classification by collapsing the ``neutral'' and ``contradiction'' classes into a single class. Distribution shift in this case comes in the form of data distribution (source data: news and various NLP datasets from pre-2005; target: crowd-sourced sentences).

\paragraph{Part-of-Speech Tagging (POS)} The POS tagging task is a sequence tagging task, where the goal is to predict the correct part-of-speech for each token in a sentence. For training data, we use the Gimpel dataset from~\cite{DBLP:conf/acl/GimpelSODMEHYFS11} with the crowd-sourced labels provided by~\cite{DBLP:conf/acl/HovyPS14} mapped to the universal POS tag set in~\cite{DBLP:conf/acl/PlankHS14}. The dataset consists of 1000 tweets (17,503 tokens) labeled with Universal POS tags and annotated by 177 annotators. Each token received at least 5 annotations. The IAA is 0.725 and the average annotator accuracy with respect to the gold labels is 67.81\%. We use the publicly available sample of the Penn Treebank POS dataset~\cite{DBLP:journals/coling/MarcusSM94} accessed from NLTK~\cite{DBLP:conf/acl/Bird06} as our out-of-domain test set, which consists of 3,914 sentences from Wall Street Journal articles (100,676 tokens). Distribution shift on this task is based on the data distribution (source: tweets, target: news).

\paragraph{Toxicity Detection} To measure performance on a highly subjective task, we use the \href{https://www.kaggle.com/competitions/jigsaw-unintended-bias-in-toxicity-classification}{toxicity detection dataset} created as a part of the Google Jigsaw unintended bias in toxicity classification competition.
The dataset we use comes from~\cite{DBLP:journals/corr/abs-2205-00501}, which annotated 25,500 comments from the original Civil Comments dataset. The pool of annotators is specifically selected and split into multiple rating pools based on self-indicated identity group membership. As this is a highly subjective task, the IAA in terms of Krippendorff's $\alpha$ is 0.196. %
We randomly split the dataset into training and test, and for the test data we use the annotations in the original crowd-sourcing task; in other words, using a completely separate annotator pool that isn't selected based on identity groups. The source of distribution shift for this task comes from the annotator pools (i.e., the data is the same but the annotators change; source: unrestricted annotator pool, target only self-identifying African American and LGBTQ annotators):

\paragraph{Image Classification} The training dataset for image classification comes from the 10,000 images in the CIFAR10 test set re-annotated by crowd workers on Amazon Mecahnical Turk (AMT)~\cite{DBLP:conf/iccv/PetersonBGR19}. The final dataset contains 511,400 annotations, with 51 annotations on average per image, and an IAA of 0.91 Krippendorff's $\alpha$. For the test data, we use the CINIC10 dataset~\cite{DBLP:journals/corr/abs-1810-03505}, which contains 210,000 images from Imagenet rescaled down to 32x32 to match the CIFAR10 image size. Here, distribution shift comes from the data distribution (source: CIFAR10 images, target: scaled down Imagenet images).

\subsection*{Significance Tests}
\begin{figure}[h]
     \centering
     \includegraphics[width=0.95\linewidth]{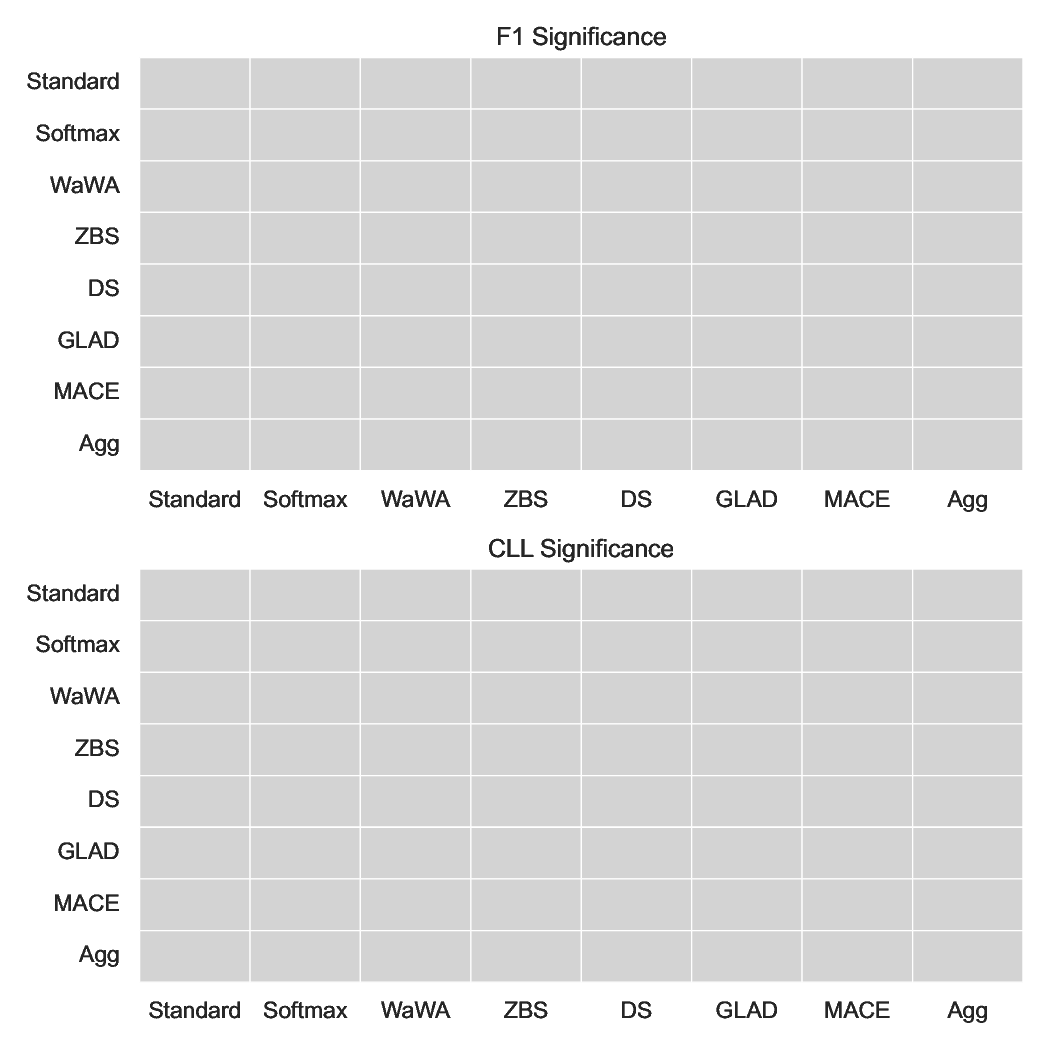}
     \caption{Significance testing for the RTE task. We apply the Bonferroni correction across the total tests (N = 56). Green indicates the method in the row is significantly better than the method in the column. Red indicates the method in the row is significantly worse than the method in the column. Grey indicates no statistically significant difference. }
     \label{fig:rte-sig-cons}
\end{figure}

\begin{figure}[h]
     \centering
     \includegraphics[width=0.95\linewidth]{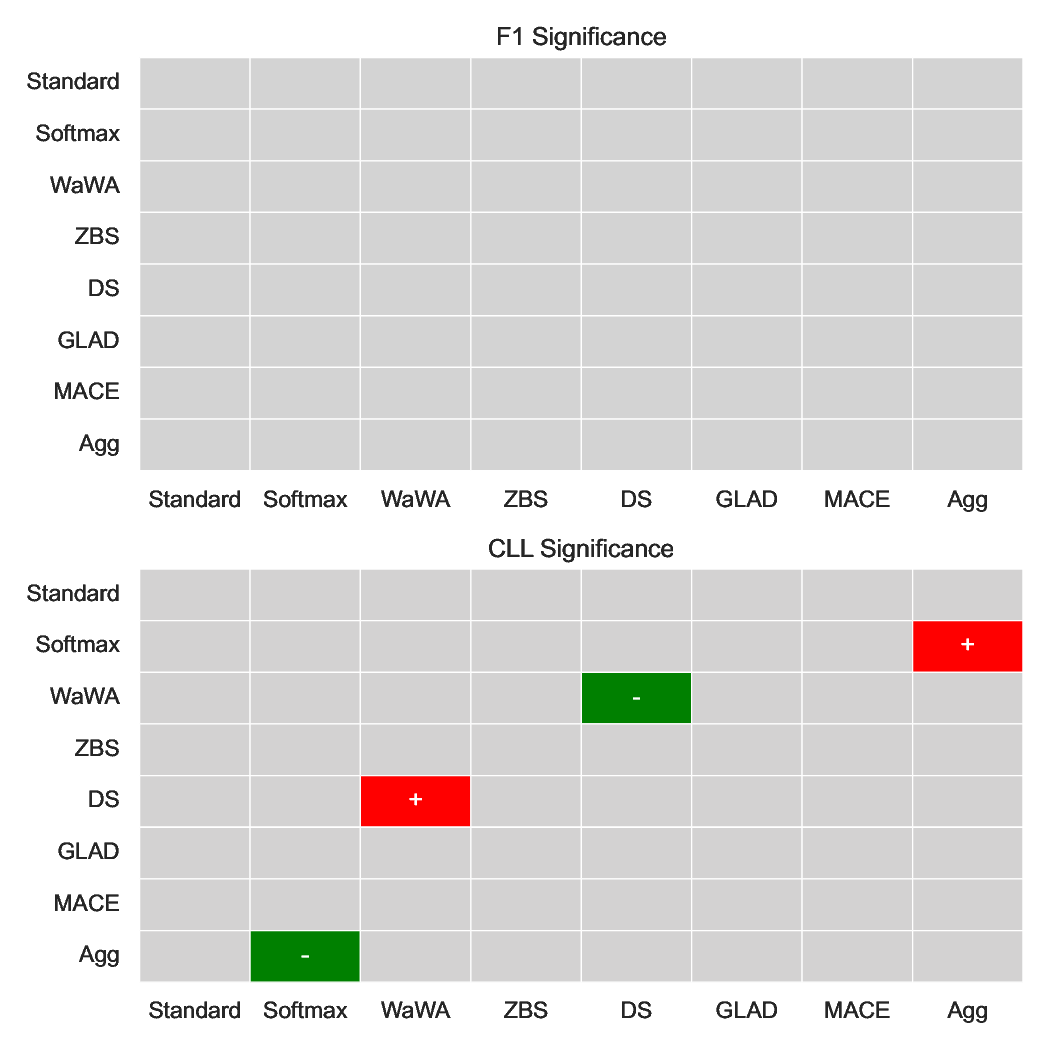}
     \caption{Significance testing for the POS task. We apply the Bonferroni correction across the total tests (N = 56). Green indicates the method in the row is significantly better than the method in the column. Red indicates the method in the row is significantly worse than the method in the column. Grey indicates no statistically significant difference. }
     \label{fig:pos-sig-cons}
\end{figure}

\begin{figure}[h]
     \centering
     \includegraphics[width=0.95\linewidth]{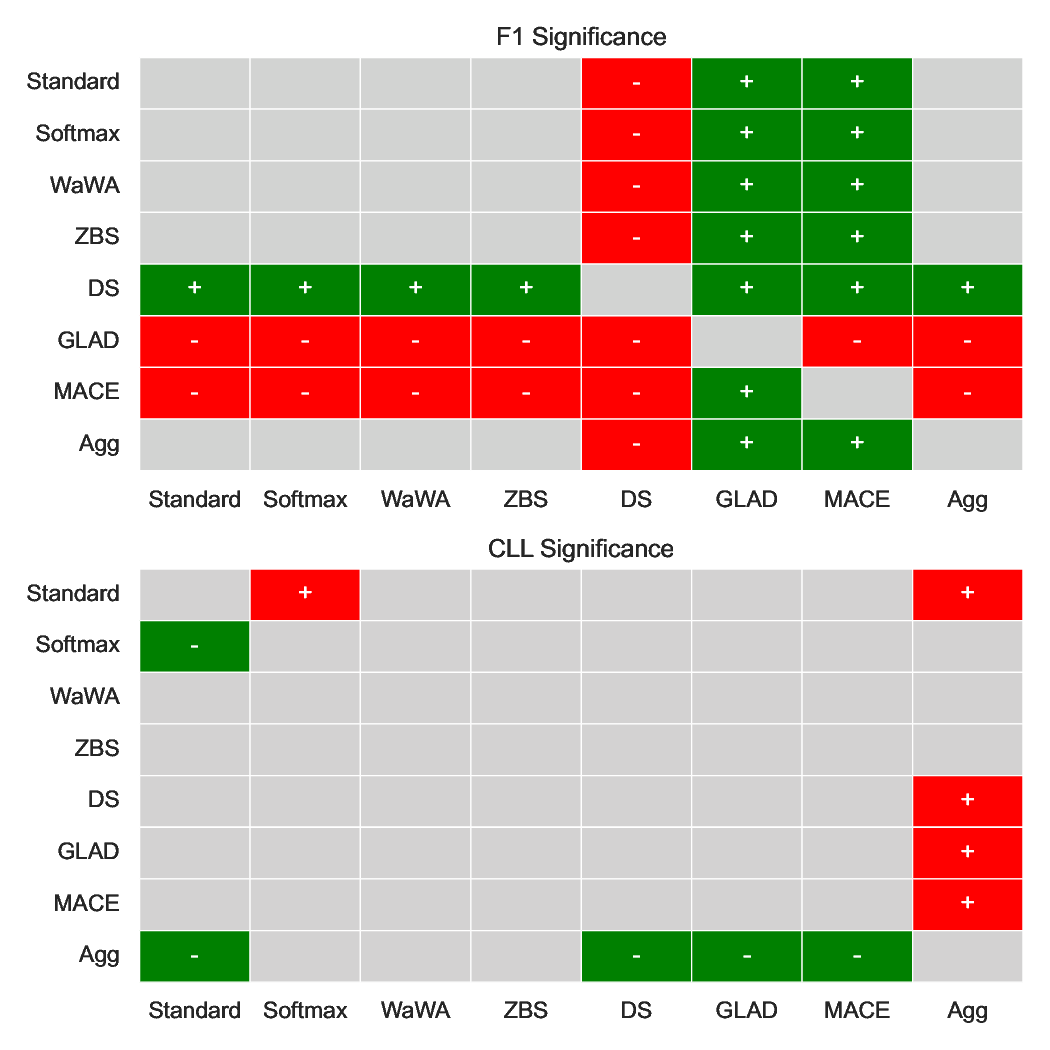}
     \caption{Significance testing for the Toxicity task. We apply the Bonferroni correction across the total tests (N = 56). Green indicates the method in the row is significantly better than the method in the column. Red indicates the method in the row is significantly worse than the method in the column. Grey indicates no statistically significant difference. }
     \label{fig:toxicity-sig-cons}
\end{figure}

\begin{figure}[h]
     \centering
     \includegraphics[width=0.95\linewidth]{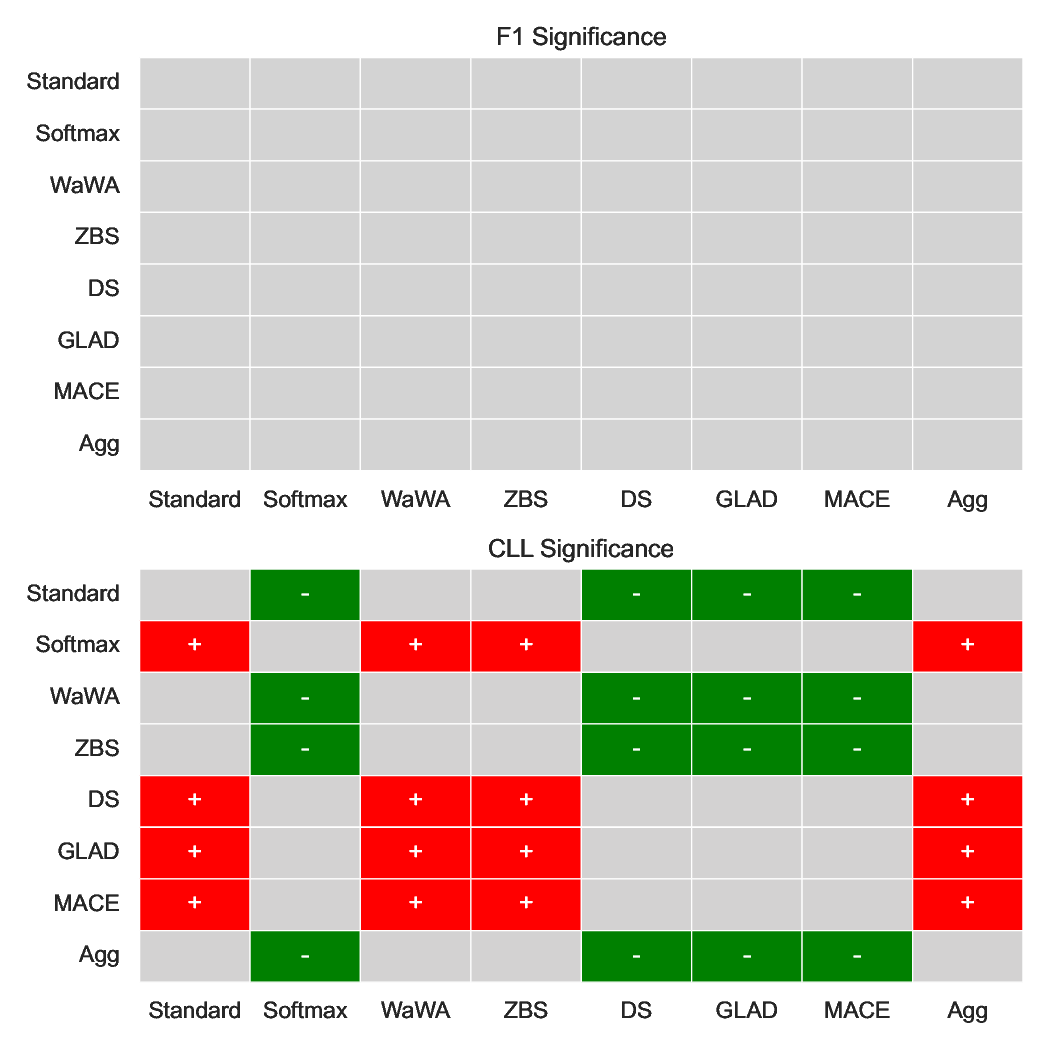}
     \caption{Significance testing for the Image Cls. task. We apply the Bonferroni correction across the total tests (N = 56). Green indicates the method in the row is significantly better than the method in the column. Red indicates the method in the row is significantly worse than the method in the column. Grey indicates no statistically significant difference. }
     \label{fig:img-sig-cons}
\end{figure}

We show a more conservative test of significance, applying the Bonferroni correction across the number of tests (N = 56) in \autoref{fig:rte-sig-cons} -- \autoref{fig:img-sig-cons}. The differences from the less conservative significance test are: for POS tagging, no method is statistically significantly different for raw performance, and the only significance in uncertainty estimation are aggregation being statistically significantly better than Softmax, and WaWA being statistically significantly better than DS. On toxicity detection, for raw performance, softmax is no longer statistically significantly better than WaWA and ZBS, and for uncertainty estimation, standard is no longer statistically significantly worse than DS, GLAD, and MACE, softmax is no longer statistically significantly better than DS, GLAD, and MACE, and ZBS is no longer statistically significantly worse than Aggregation. For RTE and image classification there is no change.

\subsection*{Evaluation Metrics}
\label{sec:eval_metrics}
\paragraph{F1}
We used the sklearn implementation of precision\_recall\_fscore \_support for F1 score, which can be found here: \url{https://scikit-learn.org/stable/modules/generated/sklearn.metrics.precision_recall_fscore_support.html}. Briefly:
\begin{equation*}
   p = \frac{tp}{tp + fp} 
\end{equation*}
\begin{equation*}
   r = \frac{tp}{tp + fn} 
\end{equation*}
\begin{equation*}
   F1 = \frac{2*p*r}{p+r} 
\end{equation*}
where $tp$ are true positives, $fp$ are false positives, and $fn$ are false negatives.

\paragraph{Calibrated Log-Likelihood} The calibrated log-likelihood is defined in \cite{DBLP:conf/iclr/AshukhaLMV20} as a method to fairly compare uncertainty estimation between models on the same test set. The key observation is that in order to obtain a fair comparison, one must first perform temperature scaling at the optimal temperature on the classifier output for each model under comparison. Additionally, this temperature must be optimized on an in-domain validation set. The procedure to calculate the calibrated log-likelihood is:
\begin{enumerate}
    \item Split the \textbf{test set} in half, one half for validation and one half for test.
    \item Optimize a temperature parameter $T$ to minimize the average negative log-likelihood $-\frac{1}{n}\sum_{i}\log \tilde{p}(y_{i}=y^{*}_{i} | x_{i})$, where $\tilde{p}_{i} = \text{softmax}(\frac{l_{i}}{T})$ and $l_{i}$ is the logits of the classifier, on the validation half of the test set.
    \item Measure the temperature scaled log-likelihood on the test half of the test set.
\end{enumerate}
Following the suggestion from \cite{DBLP:conf/iclr/AshukhaLMV20}, we run this procedure 5 times on different splits of the test set and take the average test-half log-likelihood as the result.

\subsection*{Reproducibility}

All NLP experiments were run using the RoBERTa base model released in the HuggingFace hub (\texttt{roberta-base}, \url{https://huggingface.co/roberta-base}) which has 125M parameters. All experiments with CIFAR10H/CINIC10 were run using a vision transformer with 85.8M parameters (HuggingFace: \texttt{google/vit-base-patch16-224-in21k}). We ran our experiments on a single NVIDIA TITAN RTX with 24GB of RAM. 

\paragraph{Hyperparameters} We found that performing a hyperparameter search using the in-domain validation set for hyperparameter selection can actually hurt our final performance – we tested this by performing a hyperparameter search for the RTE dataset, finding a general drop in both F1 performance and CLL on the out of domain test set, while failing to reduce the variance. The search was performed across the following values using a Bayesian hyperparameter search for 400 steps per setting:
\begin{itemize}
  \item Learning rate:  $[$1e-5, 1e-3$]$
  \item Batch size:  \{2, 4, 8, 16, 32, 64\}
  \item Epochs: \{1, 2, 3, 4, 5, 8, 10, 15\}
  \item Warmup steps: \{0, 20, 100, 200, 500, 1000\}
\end{itemize}

As such, we use good hyperparameter settings for the class of base models used (RoBERTa), which have been shown to produce highly performant classifiers across many tasks~\cite{DBLP:journals/corr/abs-1907-11692}. We used a learning rate of 2e-5 with triangular learning rate schedule using 200 warmup steps. Models are trained for 5 epochs, using the best validation F1 for the final model. The average runtimes are: 50m00s (Toxicity), 70m00s (Image Classification), 2m28s (POS), 2m39s (RTE).

\end{document}

%% file: tables/method_accuracy.tex
\begin{table}[h]
    \centering
    \caption{The accuracy of each annotation method with respect to the expert annotations in each dataset. Aggregating maintains best or near-best accuracy across tasks.}
    \begin{tabular}{|l+c |c |c |c|}
    \hline %
    \textbf{Method} & \textbf{RTE} & \textbf{POS}& \textbf{Toxicity} & \textbf{Image Cls.} \\
    \thickhline

Standard & 91.88  & 79.85  & 78.38  & 99.22 \\
\hline
Softmax & 91.88  & 79.85  & 78.38  & 99.22 \\
\hline
Wawa & 92.12  & 79.91  & 78.45  & 99.22 \\
\hline
ZBS & 92.12  & 79.96  & 78.45  & 99.22 \\
\hline
DS & 92.75  & 78.72  & 75.60  & 99.27 \\
\hline
GLAD & 92.62  & 80.07  & 77.65  & 99.22 \\
\hline
MACE & 92.50  & 80.02  & 78.13  & 99.26 \\
\thickhline
Agg & 92.88  & 80.05  & 78.43  & 99.26 \\
\hline

    \bottomrule %

    \end{tabular}
    \label{tab:dataset_accuracy}
\end{table}

%% file: tables/method_calibration.tex
\begin{table}[h]
    \centering
    \caption{Negative log likelihood of each annotation method with respect to the expert annotations in each dataset. Individual soft-labeling methods vary between tasks, while aggregating maintains best or near-best NLL.}
    \begin{tabular}{|l + c | c | c | c |}
    \hline
    Method & RTE & POS & Toxicity & Image Cls. \\
    \thickhline
Standard & 0.355  & 1.004  & 0.462  & 0.072 \\
\hline
Softmax & 0.273  & 0.834  & 0.826  & 0.051 \\
\hline
Wawa & 0.305  & 1.002  & 0.460  & 0.069 \\
\hline
ZBS & 0.304  & 1.002  & 0.460  & 0.069 \\
\hline
DS & 0.323  & 1.325  & 0.900  & 0.059 \\
\hline
GLAD & 0.235  & 1.174  & 0.940  & 0.057 \\
\hline
MACE & 0.291  & 1.261  & 0.988  & 0.060 \\
\thickhline
Agg & 0.226  & 0.871  & 0.487  & 0.046 \\
\hline

    \end{tabular}
    
    \label{tab:dataset_calibration}
\end{table}

%% file: tables/metrics.tex
\begin{table}[h]%
\begin{adjustwidth}{-2.25in}{0in}
    
    \centering
    \caption{F1 and calibrated log likelihood. Results are averaged over 10 random seeds; standard deviation is given in the subscript. Tasks marked by * are subject to input data distribution shift while datasets marked by $\dagger$ are subject to annotator pool distribution shift. Methods marked by $\ddagger$ are those which estimate either worker skill or item difficulty. Aggregating the individual soft-labeling methods yields classifiers with consistently good uncertainty estimation (best on all text based tasks) and generally good raw performance in terms of F1 across tasks.} %
    \label{tab:results}
    \begin{tabular}{|l + c | c + c | c + c | c + c | c | }
    \hline
    & \multicolumn{2}{c+}{ RTE\textsuperscript{*} } & \multicolumn{2}{c+}{ POS\textsuperscript{*} } & \multicolumn{2}{c+}{ Toxicity\textsuperscript{$\dagger$} } & \multicolumn{2}{c|}{ Image Cls.\textsuperscript{*} }\\
    \hline
    Method & F1 & CLL & F1 & CLL & F1 & CLL & F1 & CLL\\

\thickhline
Majority&$60.73_{10.16}$& $0.596_{0.05}$& $69.84_{1.72}$& $0.712_{0.03}$& $68.27_{1.16}$& $0.459_{0.01}$& $84.37_{0.29}$& $0.527_{0.01}$\\
\hline
Standard&$\mathbf{68.33_{4.71}}$& $0.637_{0.04}$& $68.54_{1.57}$& $0.708_{0.02}$& $68.85_{0.55}$& $0.613_{0.10}$& $84.51_{0.26}$& $\mathbf{0.488_{0.01}}$\\
\hline
Softmax&$62.76_{8.41}$& $0.601_{0.06}$& $69.84_{1.28}$& $1.665_{0.53}$& $68.25_{1.07}$& $0.448_{0.01}$& $84.57_{0.15}$& $0.520_{0.01}$\\
\hline
Wawa\textsuperscript{$\ddagger$}&$64.08_{8.26}$& $0.624_{0.05}$& $69.40_{1.22}$& $0.696_{0.01}$& $67.79_{1.13}$& $0.548_{0.12}$& $84.47_{0.28}$& $0.489_{0.01}$\\
\hline
ZBS\textsuperscript{$\ddagger$}&$63.50_{9.01}$& $0.629_{0.04}$& $68.96_{1.62}$& $0.703_{0.02}$& $67.40_{0.96}$& $0.546_{0.11}$& $84.46_{0.26}$& $0.489_{0.01}$\\
\hline
DS&$61.03_{7.40}$& $0.602_{0.03}$& $\mathbf{71.18_{1.22}}$& $0.730_{0.01}$& $\mathbf{71.06_{0.64}}$& $0.472_{0.01}$& $84.52_{0.21}$& $0.522_{0.01}$\\
\hline
GLAD\textsuperscript{$\ddagger$}&$62.50_{7.36}$& $0.612_{0.07}$& $70.48_{1.08}$& $0.697_{0.02}$& $60.60_{1.82}$& $0.468_{0.01}$& $84.51_{0.27}$& $0.521_{0.01}$\\
\hline
MACE\textsuperscript{$\ddagger$}&$63.27_{5.79}$& $0.601_{0.06}$& $70.40_{1.07}$& $0.700_{0.02}$& $63.97_{1.15}$& $0.467_{0.01}$& $84.46_{0.21}$& $0.524_{0.01}$\\
\thickhline
Agg&$63.26_{8.39}$& $\mathbf{0.595_{0.06}}$& $70.08_{1.51}$& $\mathbf{0.687_{0.02}}$& $68.09_{0.91}$& $\mathbf{0.440_{0.01}}$& $\mathbf{84.68_{0.24}}$& $0.500_{0.01}$\\

    \hline %

    \end{tabular}
    
\end{adjustwidth}
\end{table}

%% file: tables/significance_comparison.tex
\begin{table}%
    \begin{adjustwidth}{-2.25in}{0in}
    \centering
    \caption{A comparison of overall significance for each method. We obtain this score by comparing each method across all datasets: if method 1 is statistically significantly better than method 2, we add 1 to its score. If it is significantly worse, we subtract 1. If there is no difference, then we add 0 to the score.} %
    \label{tab:significance_comparison}
    \begin{tabular}{|l + c | c |}
    \hline %
    Method & F1 & CLL \\
    \thickhline
Standard & 1  & 0 \\
\hline
Softmax & 1  & -7  \\
\hline
Wawa & -1  & 6 \\
\hline
ZBS & 0  & 6 \\
\hline
DS & \textbf{8}  & -6 \\
\hline
GLAD & -7  & -4 \\
\hline
MACE & -5 & -4 \\
\thickhline
Agg & 1  & \textbf{9} \\

    \hline
    \end{tabular}
\end{adjustwidth}
\end{table}